\documentclass{article} 
\usepackage{iclr2027_conference,times}

\usepackage{amsmath,amsfonts,bm}

\def\eqref#1{equation~\ref{#1}}

\def\1{\bm{1}}

\def\vg{{\bm{g}}}

\def\mK{{\bm{K}}}

\def\mO{{\bm{O}}}

\def\mQ{{\bm{Q}}}

\def\mV{{\bm{V}}}
\def\mW{{\bm{W}}}
\def\mX{{\bm{X}}}

\DeclareMathAlphabet{\mathsfit}{\encodingdefault}{\sfdefault}{m}{sl}
\SetMathAlphabet{\mathsfit}{bold}{\encodingdefault}{\sfdefault}{bx}{n}

\DeclareMathOperator*{\argmin}{arg\,min}

\usepackage{float}
\usepackage{hyperref}
\usepackage{url}

\usepackage{pgfplots}
\pgfplotsset{compat=1.18}
\usepgfplotslibrary{groupplots}
\usepackage[scaled]{helvet}
\usepackage{algorithm}
\usepackage{algorithmic}
\usepackage{listings}
\usepackage{enumitem}
\usepackage{booktabs}
\usepackage{capt-of}
\usepackage[capitalize]{cleveref}

\title{Offline Makes Online Better: Calibrated Cache Compression for Video Generation}

\author{%
Zhen Wang \\
Homenet Research \\
\texttt{zhen.w@homenet.com.au}
\And
Ning Lai \\
CarryAI \\
\texttt{ning.lai@carryai.co}
\And
Qingtao Yu\thanks{Corresponding author.} \\
Australian National University \\
\texttt{terry.yu@anu.edu.au}
}

\iclrfinalcopy 
\usepackage{xspace}
\newcommand{\method}{CALM\xspace}

\begin{document}

\maketitle
\lhead{Preprint.}

\begin{abstract}
Token merging is an effective approach to reducing KV-cache memory consumption in autoregressive video generation by clustering temporally redundant tokens and retaining only a small set of representatives. 
Existing methods, however, typically apply a uniform merging ratio across all attention layers and heads. 
We observe that different layers and heads exhibit substantially different yet jointly correlated sensitivities to token merging, and that these sensitivity patterns remain consistent across prompts. Based on this observation, we introduce CALM, a Calibrated Allocation for Layer-Head Merging that uses offline calibration to improve online cache compression. Exhaustively evaluating all possible layer-head allocations is prohibitively expensive. To enable efficient calibration, we show that attention-output error strongly correlates with end-to-end video quality and can therefore serve as a lightweight proxy for allocation risk assessment. We evaluate different allocation strategies using this proxy and construct a fixed layer-head risk map offline. Given a cache budget at inference time, CALM uses dynamic programming to determine the best allocation of merging ratios. Experiments across three different autoregressive video models demonstrate that CALM consistently outperforms uniform token merging under the same memory budget,
by reducing the LPIPS to original videos by around $21\%$
with higher video-quality scores.

\end{abstract}

\section{Introduction}\label{sec:intro}

Autoregressive video diffusion models generate the frame-chunks sequentially, keeping a key-value (KV) cache that grows with every chunk \citep{wan2025,yin2024causvid,huang2025selfforcing,zhuang2026selfgradient}.
Compressing this KV cache can improve inference efficiency by reducing both memory usage and computational cost.
Token Merge has emerged as an effective training-free approach for improving model efficiency and compressing KV caches \citep{bolya2022tome,li2024vidtome,wu2025importance,hu2026centroidkv}.
In autoregressive video generation, it merges redundant tokens along the temporal dimension while retaining only a few representatives at each spatial position.
However, existing token merging methods typically keep the same number of representatives uniformly across all transformer layers and attention heads.
We argue that this uniform strategy is suboptimal and focus on addressing this limitation to enable a more fine-grained merging strategy.

To start with, we conduct a sensitivity analysis of token merging across different transformer layers and attention heads and make three key observations.
First, different layers and heads exhibit different sensitivities to merging, as measured using LPIPS against the original unmerged video.
Second, layer and head sensitivities are jointly dependent, for example, the relative sensitivity ordering of two head indices can differ across layers.
Third, these sensitivity patterns are consistent across different prompts.
The first two patterns indicate that the layer-head merging policy cannot be reduced to a simple heuristic.
Fortunately, the third observation shows that we can obtain such a policy through offline calibration and apply it across different prompts.


These observations motivate \method, an offline-calibrated allocation of temporal KV representatives across layer--head pairs. To avoid generating videos for every candidate allocation, we measure attention-output error using dense queries and shadow KV caches, building a distortion table over candidate representative counts. An offline dynamic program minimizes the summed calibrated distortion under a fixed budget. The resulting allocation is frozen and reused across unseen prompts without retraining or per-video allocation search.

Our contributions can be summarized as follows:
\begin{itemize}[leftmargin=*,itemsep=1pt]
    \item We reveal that sensitivity to temporal KV merging varies across layers and heads, with coupled layer--head effects that remain consistent across prompts.
    \item We develop a lightweight calibration proxy based on attention-output error that predicts end-to-end per-frame LPIPS without regenerating videos.
    \item We formulate temporal KV merging as a fixed-budget layer--head allocation problem and solve it exactly with dynamic programming.
    \item We show that the resulting frozen policy outperforms uniform allocation at equal budget without per-video search during inference.
\end{itemize}


\section{Related Work}\label{sec:related}

\paragraph{Autoregressive video generation.}
Recent large-scale diffusion transformers, such as Sora \citep{brooks2024sora}, CogVideoX \citep{yang2025cogvideox}, and Wan \citep{wan2025}, have achieved remarkable success in video generation.
Nowadays, autoregressive video diffusion has emerged as a mainstream paradigm for long-form generation, producing future frames or chunks conditioned on previously generated content.
Diffusion Forcing trains causal sequence models using independently sampled noise levels for different frames \citep{chen2024diffusionforcing}.
CausVid distills a bidirectional Wan teacher into a few-step causal generator through distribution-matching distillation \citep{wan2025,yin2024dmd,yin2024causvid}.
Self-Forcing trains on self-generated rollouts to reduce the train--test discrepancy \citep{huang2025selfforcing},
while Self Gradient Forcing propagates future supervision through historical KV states \citep{zhuang2026selfgradient}.
These causal models retain past KV states during generation, making cache efficiency important. Sparse Forcing learns sparse attention \citep{xu2026sparseforcing},
whereas Forcing-KV performs head-wise cache pruning \citep{ji2026forcingkv}.
Our method instead merges redundant cache tokens and calibrates the per-(layer, head) merge budget offline.

\paragraph{Token merging and KV-cache compression.}
Token merging reduces transformer computation by combining redundant tokens into shared representatives \citep{bolya2022tome}. Attention heads also exhibit substantial specialization and differences in pruning sensitivity \citep{voita2019analyzing}. For LLMs, KV-cache eviction methods select entries by position or attention statistics: StreamingLLM retains initial attention sinks \citep{xiao2024streamingllm}, H2O retains heavy hitters \citep{zhang2023h2o}, and SnapKV uses an end-of-prompt observation window \citep{li2024snapkv}. Merge-based methods instead aggregate cached states. KVMerger clusters similar keys \citep{wang2024model}; CaM and D2O merge evicted states into retained states \citep{zhang2024cam, wan2025d2o}; MiniCache shares KV states across adjacent layers \citep{liu2024minicache}; and WeightedKV applies attention-weighted fusion \citep{yuan2025weightedkv}. Token merging has also been applied to vision transformers and diffusion models \citep{bolya2023token, liang2022not, chen2023diffrate, zeng2022not, wu2025importance}, video transformers \citep{choi2024vidtldr, li2024vidtome, choudhury2024dont}, and video diffusion transformers \citep{sun2025asymrnr, lu2025toma}. Our work studies temporal KV-cache merging with an offline-calibrated per-(layer, head) budget.


\section{Preliminaries}

\paragraph{KV Cache in Autoregressive Video Generative Models.}
A key-value (KV) cache stores previously computed attention keys and values for reuse in subsequent autoregressive steps.
Take Self-Forcing \citep{huang2025selfforcing} as an example.
Consider a causal video transformer with $L$ layers, token dimension $D$, and $H$ attention heads of dimension $d=D/H$.
For chunk $c$, let $F_c$ denote its number of latent frames and $S$ the number of spatial tokens per frame, giving $N_c=F_cS$ input tokens $\mX_l^c\in\mathbb{R}^{B\times N_c\times D}$ at layer $l$.
At each denoising step, the layer computes
\begin{equation}
    \mQ_l^c=\operatorname{RoPE}\!\left(\operatorname{Norm}_q(\mX_l^c\mW_l^Q)\right),\quad
    \mK_l^c=\operatorname{RoPE}\!\left(\operatorname{Norm}_k(\mX_l^c\mW_l^K)\right),\quad
    \mV_l^c=\mX_l^c\mW_l^V,
\end{equation}
where each tensor is reshaped to $\mathbb{R}^{B\times N_c\times H\times d}$.
Before processing chunk $c$, layer $l$ retains $\mK_l^{<c}$ and $\mV_l^{<c}$ in $\mathbb{R}^{B\times N_{<c}\times H\times d}$, where $N_{<c}=\sum_{j<c}N_j$.
The current queries attend jointly to the cached context and the current chunk:
\begin{equation}
    \mO_l^c=\operatorname{Softmax}\!\left(
    \frac{\mQ_l^c[\mK_l^{<c};\mK_l^c]^{\top}}{\sqrt d}
    \right)[\mV_l^{<c};\mV_l^c],
\end{equation}
where $[\,;\,]$ denotes concatenation along the token dimension and the head dimension is omitted for clarity.
During chunk denoising, the historical cache is reused while the current K/V slots are overwritten at each step.
After the chunk is denoised, Self-Forcing performs one additional context forward at timestep zero and stores the resulting clean K/V for subsequent chunks.
Thus, after chunk $c$, each layer holds a cache of shape $B\times N_{\leq c}\times H\times d$ for both keys and values; the Wan-1.3B configuration used by Self-Forcing has $L=30$, $H=12$, $d=128$, and $S=1560$.


\paragraph{TokenMerge for KV Cache Compression.}
In sequential token merging, a KV cache $\mK,\mV\in\mathbb{R}^{T\times S\times d}$ is compressed to at most $r$ temporal representatives independently at each spatial position.
We apply farthest-point sampling to the normalized keys to select $r$ representatives, then assign each temporal token to its most similar representative by cosine similarity.
Each cache entry carries the number $c_i$ of original tokens it represents, with $c_i=1$ for new tokens.
On each merge, we sum these counts within each cluster $G_j$, retain the selected key, and average values weighted by $c_i$.
To account for cluster size in attention, we add $\log g_j$ to each representative's logit:
\begin{equation}
    \widetilde{\mO}
    =\operatorname{Softmax}\!\left(
        \frac{\mQ\widetilde{\mK}^{\top}}{\sqrt d}
        +\log \vg^{\top}
    \right)\widetilde{\mV},
    \qquad g_j=\sum_{i\in G_j}c_i,
    \label{eq:cluster-corrected-attention}
\end{equation}
where $\log \vg^{\top}$ is broadcast across queries.
\Cref{alg:temporal-kv-merge} summarizes the procedure.

\begin{algorithm}[t]
\caption{Temporal KV Token Merging}
\label{alg:temporal-kv-merge}
\definecolor{codecomment}{rgb}{0.10,0.55,0.10}
\definecolor{codekeyword}{rgb}{0.70,0.12,0.45}
\lstset{
    backgroundcolor=\color{white},
    basicstyle=\fontsize{7pt}{7.5pt}\ttfamily\selectfont,
    columns=fullflexible,
    breaklines=true,
    commentstyle=\color{codecomment},
    keywordstyle=\color{codekeyword},
    showstringspaces=false
}
\begin{lstlisting}[language=Python]
def temporal_kv_merge(K, V, count, R):
    # K, V: [T, S, d]; count: [T, S], ones for new tokens
    # R: representatives; attention adds log(count) to logits
    T, S, d = K.shape
    if T <= R:
        return K, V, count  # keep existing counts

    K_out = K.new_empty(R, S, d)
    V_out = V.new_empty(R, S, d)
    count_out = count.new_empty(R, S)

    for s in range(S):
        K_norm = normalize(K[:, s], dim=-1)
        rep = farthest_point_sample(K_norm, R)
        sim = K_norm @ K_norm[rep].T
        group = sim.argmax(dim=-1)
        group[rep] = arange(R)  # each representative owns itself

        for j in range(R):
            members = group == j
            w = count[members, s]
            K_out[j, s] = K[rep[j], s]
            V_out[j, s] = (w[:, None] * V[members, s]).sum(0) / w.sum()
            count_out[j, s] = w.sum()

    return K_out, V_out, count_out
\end{lstlisting}
\end{algorithm}


\section{Why Offline Could Help Online?}

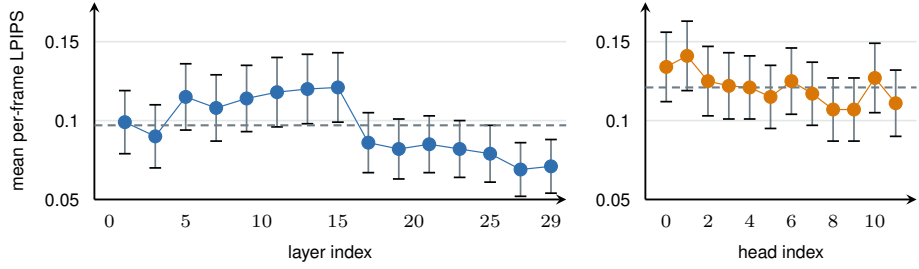
\begin{figure}[t]
    \centering
    \definecolor{obsrust}{HTML}{BF4A25}
    \definecolor{obsblue}{HTML}{2F6FB0}
    \definecolor{obsorange}{HTML}{D97706}
    \definecolor{obsgrey}{HTML}{6F7D86}
    \definecolor{obsgrid}{HTML}{E2E6E5}
    \definecolor{obsrule}{HTML}{C6CECD}
    \hspace*{-0.06\linewidth}
    \begin{tikzpicture}
        \begin{groupplot}[
            group style={group size=2 by 1,horizontal sep=0.075\linewidth},
            height=0.30\linewidth,
            ymin=0.05, ymax=0.175,
            ytick={0.05,0.10,0.15},
            yticklabels={0.05,0.1,0.15},
            yticklabel style={font=\sffamily\scriptsize},
            xticklabel style={font=\sffamily\scriptsize},
            xlabel style={font=\sffamily\scriptsize},
            ylabel style={font=\sffamily\scriptsize},
            axis line style={draw=black,line width=0.7pt},
            tick style={draw=none},
            axis lines=left,
            ymajorgrids=true,
            grid style={draw=obsgrid,line width=0.6pt},
            error bars/y dir=both,
            error bars/y explicit,
            error bars/error bar style={draw=obsgrey,line width=0.65pt},
            error bars/error mark=-,
            error bars/error mark options={rotate=90,mark size=2.2pt,line width=0.65pt},
        ]
        \nextgroupplot[
            width=0.56\linewidth,
            xlabel={layer index},
            ylabel={mean per-frame LPIPS},
            xmin=-1, xmax=30,
            xtick={0,5,10,15,20,25,29},
        ]
        \addplot[
            mark=*,mark size=2.4pt,draw=obsblue,
            mark options={fill=obsblue},line width=0.45pt,
        ] table[x=layer,y=mean,y error=err] {
            layer mean err
            1  0.099 0.020
            3  0.090 0.020
            5  0.115 0.021
            7  0.108 0.021
            9  0.114 0.021
            11 0.118 0.022
            13 0.120 0.022
            15 0.121 0.022
            17 0.086 0.019
            19 0.082 0.019
            21 0.085 0.018
            23 0.082 0.018
            25 0.079 0.018
            27 0.069 0.017
            29 0.071 0.017
        };
        \addplot[densely dashed,draw=obsgrey,line width=0.8pt,mark=none]
            coordinates {(-1,0.097) (30,0.097)};

        \nextgroupplot[
            width=0.37\linewidth,
            xlabel={head index},
            xmin=-1, xmax=12,
            xtick={0,2,4,6,8,10},
        ]
        \addplot[
            mark=*,mark size=2.4pt,draw=obsorange,
            mark options={fill=obsorange},line width=0.45pt,
        ] table[x=head,y=mean,y error=err] {
            head mean err
            0  0.134 0.022
            1  0.141 0.022
            2  0.125 0.022
            3  0.122 0.021
            4  0.121 0.020
            5  0.115 0.020
            6  0.125 0.021
            7  0.117 0.020
            8  0.107 0.020
            9  0.107 0.020
            10 0.127 0.022
            11 0.111 0.021
        };
        \addplot[densely dashed,draw=obsgrey,line width=0.8pt,mark=none]
            coordinates {(-1,0.121) (12,0.121)};
        \end{groupplot}
    \end{tikzpicture}
    \vspace{-4mm}
    \caption{Mean per-frame LPIPS relative to the unmerged video with $R{=}6$.
    \textbf{(left)} All heads within each evaluated layer are merged; for clarity, we show every second layer.
    \textbf{(right)} Each head is merged across all layers.
    Robustness to merging varies across both layers and heads.}
    \label{fig:obs-sensitivity}
\end{figure}

\paragraph{Different layers and heads have different sensitivity.}
To determine whether sensitivity to temporal merging differs across layers and heads, we conduct an analysis along both axes.
At a fixed merge budget of $R=6$, each run merges only one unit: either all 12 heads in one layer, yielding 30 layer configurations;
or one head index across all 30 layers, yielding 12 head configurations.
We evaluate each configuration on 100 prompts using mean per-frame LPIPS against the unmerged video.
Layer sensitivity varies substantially: layer~0 is among the most fragile at LPIPS~$0.12$, whereas layer~27 is the most robust at $0.069$.
Layer~18 is a notable exception to this overall trend, reaching $0.148$ and producing a $0.079$ range across layers.
Head sensitivity also varies, with mean LPIPS ranging from $0.141$ for head~1 to $0.107$ for head~8, a spread of $0.034$.
Compared with layers, the variation across heads is less pronounced but still worth noticing.
\Cref{fig:obs-sensitivity} presents both sweeps.


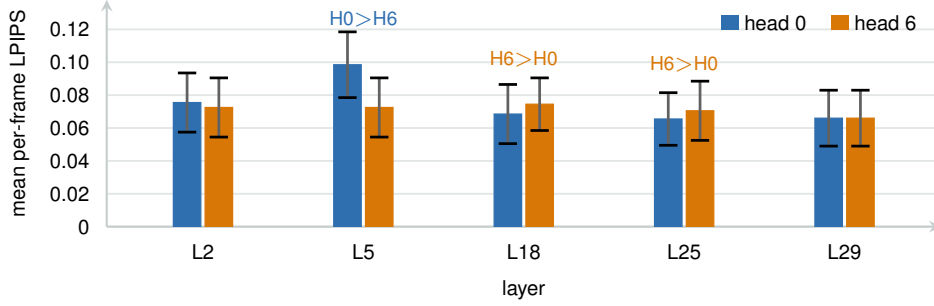
\begin{figure}[t]
    \centering
    \definecolor{obsblue}{HTML}{2F6FB0}
    \definecolor{obsorange}{HTML}{D97706}
    \definecolor{obsgrey}{HTML}{6F7D86}
    \definecolor{obsgrid}{HTML}{E2E6E5}
    \definecolor{obsrule}{HTML}{C6CECD}
    \begin{tikzpicture}[scale=1.15,transform shape]
        \begin{axis}[
            width=0.80\linewidth,
            height=0.3\linewidth,
            ybar=1.4pt,
            bar width=9pt,
            xmin=0.4,xmax=5.6,
            xtick={1,2,3,4,5},
            xticklabels={L2,L5,L18,L25,L29},
            xlabel={layer},
            ymin=0,ymax=0.138,
            ytick={0,0.02,0.04,0.06,0.08,0.10,0.12},
            yticklabels={0,0.02,0.04,0.06,0.08,0.10,0.12},
            ylabel={mean per-frame LPIPS},
            yticklabel style={font=\sffamily\scriptsize},
            xticklabel style={font=\sffamily\scriptsize},
            xlabel style={font=\sffamily\scriptsize},
            ylabel style={font=\sffamily\scriptsize},
            axis line style={draw=obsrule,line width=0.7pt},
            tick style={draw=none},
            axis lines=left,
            ymajorgrids=true,
            grid style={draw=obsgrid,line width=0.6pt},
            error bars/y dir=both,error bars/y explicit,
            error bars/error bar style={draw=black!62,line width=0.9pt},
            error bars/error mark=-,
            error bars/error mark options={rotate=90,mark size=3pt,line width=0.9pt},
            legend style={font=\sffamily\scriptsize,draw=none,fill=none,
                at={(0.99,0.99)},anchor=north east,legend columns=2,/tikz/every even column/.append style={column sep=6pt}},
            legend image code/.code={\draw[#1,draw=none] (0cm,-0.06cm) rectangle (0.20cm,0.11cm);},
        ]
        \addplot[fill=obsblue,draw=obsblue] table[x=L,y=v,y error=e] {
            L v      e
            1 0.0755 0.018
            2 0.0985 0.020
            3 0.0685 0.018
            4 0.0655 0.016
            5 0.0660 0.017
        };
        \addplot[fill=obsorange,draw=obsorange] table[x=L,y=v,y error=e] {
            L v      e
            1 0.0725 0.018
            2 0.0725 0.018
            3 0.0745 0.016
            4 0.0705 0.018
            5 0.0660 0.017
        };
        \legend{head 0,head 6}
        \node[font=\sffamily\fontsize{6.6}{7}\selectfont,color=obsblue,anchor=south] at (axis cs:2,0.115) {H0$>$H6};
        \node[font=\sffamily\fontsize{6.6}{7}\selectfont,color=obsorange,anchor=south] at (axis cs:3,0.091) {H6$>$H0};
        \node[font=\sffamily\fontsize{6.6}{7}\selectfont,color=obsorange,anchor=south] at (axis cs:4,0.088) {H6$>$H0};
        \end{axis}
    \end{tikzpicture}
    \vspace{-4mm}
    \caption{Sensitivity of two head indices across selected layers when one layer and head pair is merged.
    Points report mean LPIPS, and error bars report confidence intervals.
    The sensitivity ordering reverses across layers, showing that layer and head effects are jointly dependent.}
    \label{fig:obs-joint}
\end{figure}

\paragraph{Layer and head sensitivities are jointly dependent.}
We next examine whether layer and head sensitivities are jointly dependent rather than separable across the two axes.
This distinction is important because, if they were separable, one could estimate their effects independently and combine them into a simple allocation heuristic.
We disprove the independent assumption by identifying counterexamples.
For each layer, we merge the same two head indices, 0 and 6, one at a time and compare their sensitivities.
For example, in layer~5, head~0 is more sensitive than head~6, with LPIPS $0.098$ versus $0.072$.
The ordering reverses in layers~18 and~25, where head~6 is more sensitive, with LPIPS $0.074$ versus $0.069$ and $0.070$ versus $0.065$, respectively.
Thus, the effects of merging are interdependent and complex across layers and heads, motivating us to calibrate the entire allocation map directly.
\Cref{fig:obs-joint} illustrates this joint dependency.


\paragraph{The patterns are consistent across prompts.}
Finally, we examine whether the same layers and heads exhibit similar sensitivity patterns across different prompts.
We split the 100 prompts equally into calibration and test sets, obtain layer-wise and head-wise sensitivity rankings from the calibration set, and measure their Spearman correlation with the rankings of individual test prompts.
We use a shuffled calibration ranking as the control.
The rankings remain consistent across prompts, with median correlations of $0.69$ for layers and $0.41$ for heads, compared with approximately $0$ for the shuffled control.
This consistency allows us to estimate layer and head sensitivities in advance through offline calibration and use them to allocate the token budget more effectively during online token merging on unseen prompts.
\Cref{fig:obs-consistency} reports the per-prompt correlations.


\begin{figure*}[t]
    \centering
    \definecolor{obsblue}{HTML}{2F6FB0}
    \definecolor{obsorange}{HTML}{D97706}
    \definecolor{obsgrid}{HTML}{E2E6E5}
    \definecolor{obsrule}{HTML}{C6CECD}
    \begin{minipage}[t]{0.55\linewidth}
        \centering
        \begin{minipage}[t]{\linewidth}
        \centering
        \begin{tikzpicture}
            \begin{axis}[
                width=1.0\linewidth,height=0.192\textheight,
                xmin=-0.1,xmax=1.05,ymin=0.40,ymax=2.60,
                xlabel={test-prompt Spearman vs.\ calibration ranking},
                xlabel style={font=\sffamily\scriptsize},
                xtick={-1,-0.5,0,0.5,1},xticklabel style={font=\sffamily\scriptsize},
                ytick={1,2},yticklabels={head,layer},
                yticklabel style={font=\sffamily\scriptsize},
                axis lines=left,axis line style={draw=obsrule,line width=0.7pt},tick style={draw=none},
                xmajorgrids=true,grid style={draw=obsgrid,line width=0.6pt},clip=false,
            ]
            \draw[densely dashed,draw=obsrule,line width=0.8pt] (axis cs:0,0.45) -- (axis cs:0,2.55);
            \draw[obsblue,line width=0.8pt] (axis cs:0.26,2)--(axis cs:0.582,2);
            \draw[obsblue,line width=0.8pt] (axis cs:0.798,2)--(axis cs:0.90,2);
            \draw[obsblue,line width=0.8pt] (axis cs:0.26,1.88)--(axis cs:0.26,2.12);
            \draw[obsblue,line width=0.8pt] (axis cs:0.90,1.88)--(axis cs:0.90,2.12);
            \fill[obsblue,fill opacity=0.16] (axis cs:0.582,1.74) rectangle (axis cs:0.798,2.26);
            \draw[obsblue,line width=0.8pt] (axis cs:0.582,1.74) rectangle (axis cs:0.798,2.26);
            \draw[obsblue,line width=1.5pt] (axis cs:0.695,1.74)--(axis cs:0.695,2.26);
            \draw[obsblue,line width=0.8pt] (axis cs:0.08,1)--(axis cs:0.363,1);
            \draw[obsblue,line width=0.8pt] (axis cs:0.577,1)--(axis cs:0.84,1);
            \draw[obsblue,line width=0.8pt] (axis cs:0.08,0.88)--(axis cs:0.08,1.12);
            \draw[obsblue,line width=0.8pt] (axis cs:0.84,0.88)--(axis cs:0.84,1.12);
            \fill[obsblue,fill opacity=0.16] (axis cs:0.363,0.74) rectangle (axis cs:0.577,1.26);
            \draw[obsblue,line width=0.8pt] (axis cs:0.363,0.74) rectangle (axis cs:0.577,1.26);
            \draw[obsblue,line width=1.5pt] (axis cs:0.415,0.74)--(axis cs:0.415,1.26);
            \node[anchor=south,font=\sffamily\fontsize{6.8}{7}\selectfont,text=obsblue] at (axis cs:0.695,2.30) {median 0.69};
            \node[anchor=south,font=\sffamily\fontsize{6.8}{7}\selectfont,text=obsblue] at (axis cs:0.415,1.30) {median 0.41};
            \end{axis}
        \end{tikzpicture}
        \end{minipage}
        \vspace{-8mm}
        \captionof{figure}{Sensitivity consistency across prompts.
        Boxes show the correlation between each test prompt and the calibration ranking, with quartiles reported for layers and heads.
        Both rankings remain consistent across prompts.}
        \label{fig:obs-consistency}
    \end{minipage}\hfill
    \begin{minipage}[t]{0.43\linewidth}
        \centering
        \begin{minipage}[t]{\linewidth}
        \centering
        \begin{tikzpicture}
            \begin{axis}[
                width=1.0\linewidth,height=0.192\textheight,
                xmode=log,ymode=log,xmin=3e-6,xmax=0.13,ymin=8e-5,ymax=0.9,
                xlabel={attention-output error (rel. MSE)},ylabel={video LPIPS},
                yticklabel style={font=\sffamily\scriptsize},xticklabel style={font=\sffamily\scriptsize},
                xlabel style={font=\sffamily\scriptsize},ylabel style={font=\sffamily\scriptsize},
                axis line style={draw=obsrule,line width=0.7pt},tick style={draw=none},axis lines=left,
                ymajorgrids=true,xmajorgrids=true,grid style={draw=obsgrid,line width=0.6pt},
            ]
            \addplot[only marks,mark=*,mark size=1.0pt,draw=none,fill=obsblue,fill opacity=0.32]
                table {figures/assets/proxy_scatter.dat};
            \addplot[domain=5e-6:0.1,samples=60,draw=obsorange,line width=1.4pt,no marks]{8.236*x^(0.911)};
            \node[font=\sffamily\scriptsize,anchor=north west,align=left,text=obsorange]
                at (rel axis cs:0.03,0.97) {$r=0.94$ (log--log)\\[-1pt]$\rho=0.90$};
            \end{axis}
        \end{tikzpicture}
        \end{minipage}
        \vspace{-8mm}
        \captionof{figure}{Validation of attention output error as a calibration proxy.
        The plot compares attention output error with video LPIPS across representative counts.
        Their strong correlation shows that the proxy predicts video distortion.}
        \label{fig:proxy}
    \end{minipage}
\end{figure*}

\section{Offline Calibration for Online Merging}

CALM builds a distortion table $D\in\mathbb{R}^{L\times H\times|\mathcal{R}|}$, where $D_{l,h}(r)$ measures attention-output error when a layer--head pair retains $r$ temporal representatives. We use this lightweight proxy instead of generating videos for every candidate allocation, then solve a fixed-budget allocation problem offline and freeze the resulting map for inference.


\paragraph{Attention relative MSE is the right proxy.}
We first verify that attention output error is strongly correlated with final video LPIPS and can therefore serve as a faithful calibration proxy.
For each layer $l$, head $h$, and representative count $r$, we directly compute the attention output relative MSE as
\begin{equation}
    D^{\mathrm{raw}}_{l,h}(r)
    =\mathbb{E}_{x}\!\left[
        \frac{
            \lVert \mO^{(r)}_{l,h}(x)-\mO^{\mathrm{full}}_{l,h}(x)\rVert_F^2
        }{
            \lVert \mO^{\mathrm{full}}_{l,h}(x)\rVert_F^2
        }
    \right],\qquad D^{\mathrm{raw}}\in\mathbb{R}^{L\times H\times|\mathcal{R}|},
    \label{eq:relmse}
\end{equation}
where the norms span the full rollout, including denoising and clean-context passes, and ratios are averaged over prompts.
Across $17$ representative counts on each of $40$ prompts, attention-output error correlates with video LPIPS (Spearman $\rho{=}0.90$; log-space Pearson $r{=}0.94$), supporting its use as a calibration proxy (\cref{fig:proxy,app:proxy}).

\paragraph{Risk map calibration.}
We build the distortion table from dense-generation rollouts.
For each prompt and $r\in\mathcal{R}$, dense queries attend to a shadow cache recursively merged from clean K/V, plus the current chunk's K/V; we compare outputs with dense attention.
Pooling squared errors and dense-output energies over each rollout, then averaging per-prompt ratios, gives $D^{\mathrm{raw}}_{l,h}(r)$.
Each $D_{l,h}(r)$ estimates the distortion from merging head $h$ in layer $l$ to $r$ representatives.

\begin{algorithm}[t]
\caption{Distortion-table calibration}
\label{alg:calibrate}
\definecolor{codecomment}{rgb}{0.10,0.55,0.10}
\definecolor{codekeyword}{rgb}{0.70,0.12,0.45}
\lstset{
    backgroundcolor=\color{white},
    basicstyle=\fontsize{7pt}{7.5pt}\ttfamily\selectfont,
    columns=fullflexible,
    breaklines=true,
    commentstyle=\color{codecomment},
    keywordstyle=\color{codekeyword},
    showstringspaces=false
}
\begin{lstlisting}[language=Python]
# Input: model; prompts C with fixed seeds; increasing grid R
# Output: processed distortion table D with shape [L, H, |R|]
# sum_sq reduces batch, token, channel axes, retaining heads
sq_err, sq_full = zeros(|C|, L, H, |R|), zeros(|C|, L, H, |R|)
for x_id, x in enumerate(C):
    for r_id, r in enumerate(R):
        shadow_cache = [empty_cache() for l in range(L)]  # K, V, count
        for b in dense_rollout(model, x, seed=seed(x)):  # merging off
            for p in passes:  # 4 denoising steps + clean-context pass
                for l in range(L):
                    Q, K_b, V_b, O_full = dense_attention(model, x, b, p, l)
                    K, V, count = concat(shadow_cache[l], (K_b, V_b, ones))
                    O_r = attention(Q, K, V, bias=log(count))
                    sq_err[x_id, l, :, r_id] += sum_sq(O_r - O_full)
                    sq_full[x_id, l, :, r_id] += sum_sq(O_full)
            for l in range(L):  # clean K/V of chunk b persist
                K_b, V_b = clean_kv(b, l)
                K, V, count = concat(shadow_cache[l], (K_b, V_b, ones))
                shadow_cache[l] = temporal_kv_merge(K, V, count, r)  # per head
D_raw = mean(sq_err / sq_full, dim=0)  # per-prompt ratio, then mean
D = cummin(D_raw, dim=-1)  # running minimum along R
return D
\end{lstlisting}
\end{algorithm}

\paragraph{Budget Allocation with Dynamic Programming.}
Given the distortion map above, we aim to obtain the optimal allocation under the given budget $B$ before inference.
Formally, we solve
\begin{equation}
P^\star=\argmin_{P}\sum_{l,h}D_{l,h}(P_{l,h})
\quad\text{s.t.}\quad \sum_{l,h}P_{l,h}=B,\qquad P_{l,h}\in\mathcal{R}_{\mathrm{search}},
\label{eq:alloc}
\end{equation}
where $D$ is the processed table and $\mathcal{R}_{\mathrm{search}}=\{r\in\mathcal{R}:r_{\min}\le r\le r_{\max}\}$ keeps the measured counts within the bounds.
We solve this discrete allocation problem exactly with dynamic programming.
The resulting map $P$ minimizes the predicted distortion under the given budget, as shown in \cref{alg:allocate}.
We also require every unit to retain at least $r_{\min}$ representatives.
This lower bound prevents overly aggressive compression, which otherwise degrades the early frames, as analyzed in \cref{sec:floor-analysis}.

\begin{figure*}[t]
    \centering
    \definecolor{obsblue}{HTML}{2F6FB0}
    \definecolor{obsorange}{HTML}{D97706}
    \definecolor{obsgrey}{HTML}{6F7D86}
    \definecolor{obsgrid}{HTML}{E2E6E5}
    \definecolor{obsrule}{HTML}{C6CECD}
    \begin{minipage}[t]{0.49\linewidth}
        \centering
        \begin{tikzpicture}
            \begin{axis}[
                width=1.0\linewidth,height=0.62\linewidth,
                xmin=0,xmax=177,xtick={0,30,60,90,120,150},
                xlabel={frame index},ymin=0,ymax=0.52,ytick={0,0.1,0.2,0.3,0.4,0.5},
                ylabel={per-frame LPIPS},
                yticklabel style={font=\sffamily\scriptsize},xticklabel style={font=\sffamily\scriptsize},
                xlabel style={font=\sffamily\scriptsize},ylabel style={font=\sffamily\scriptsize},
                axis line style={draw=obsrule,line width=0.7pt},tick style={draw=none},axis lines=left,
                ymajorgrids=true,grid style={draw=obsgrid,line width=0.6pt},
                legend style={font=\sffamily\scriptsize,draw=none,fill=none,at={(0.02,0.98)},anchor=north west,
                    legend cell align=left,row sep=-1pt},
            ]
            \addplot[obsgrey,line width=0.9pt,no marks] table[x=frame,y=uniform] {figures/assets/floor_curves.dat};
            \addplot[obsorange,line width=0.9pt,no marks] table[x=frame,y=nofloor] {figures/assets/floor_curves.dat};
            \addplot[obsblue,line width=1.4pt,no marks] table[x=frame,y=rmin6] {figures/assets/floor_curves.dat};
            \legend{uniform $R{=}8$, \method\ ($r_{\min}{=}2$), \method\ ($r_{\min}{=}6$)}
            \end{axis}
        \end{tikzpicture}\\[1pt]
        {\footnotesize\sffamily (a) per-frame LPIPS ($\downarrow$)}
    \end{minipage}\hfill
    \begin{minipage}[t]{0.49\linewidth}
        \centering
        \begin{tikzpicture}
            \begin{axis}[
                width=1.0\linewidth,height=0.62\linewidth,
                xmin=0,xmax=177,xtick={0,30,60,90,120,150},
                xlabel={frame index},ymin=15,ymax=53,ytick={20,30,40,50},
                ylabel={per-frame PSNR (dB)},
                yticklabel style={font=\sffamily\scriptsize},xticklabel style={font=\sffamily\scriptsize},
                xlabel style={font=\sffamily\scriptsize},ylabel style={font=\sffamily\scriptsize},
                axis line style={draw=obsrule,line width=0.7pt},tick style={draw=none},axis lines=left,
                ymajorgrids=true,grid style={draw=obsgrid,line width=0.6pt},
            ]
            \addplot[obsgrey,line width=0.9pt,no marks] table[x=frame,y=uniform] {figures/assets/floor_psnr.dat};
            \addplot[obsorange,line width=0.9pt,no marks] table[x=frame,y=nofloor] {figures/assets/floor_psnr.dat};
            \addplot[obsblue,line width=1.4pt,no marks] table[x=frame,y=rmin6] {figures/assets/floor_psnr.dat};
            \end{axis}
        \end{tikzpicture}\\[1pt]
        {\footnotesize\sffamily (b) per-frame PSNR ($\uparrow$)}
    \end{minipage}
    \vspace{-2mm}
    \caption{Temporal quality under a fixed cache budget relative to the unmerged video.
    Panel (a) reports per frame LPIPS, and panel (b) reports per frame PSNR.
    The curves compare uniform allocation, calibrated allocation at the grid-minimum floor ($r_{\min}{=}2$), and calibrated allocation with $r_{\min}{=}6$.
    The lower bound preserves early frame quality while calibrated allocation improves later frames.}
    \label{fig:floor}
\end{figure*}
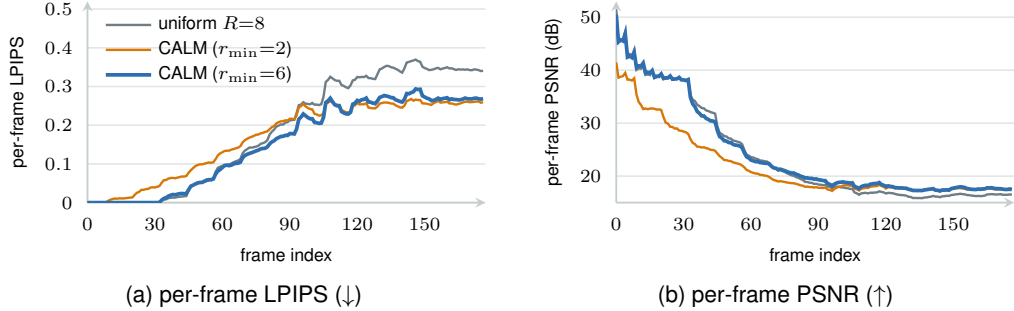

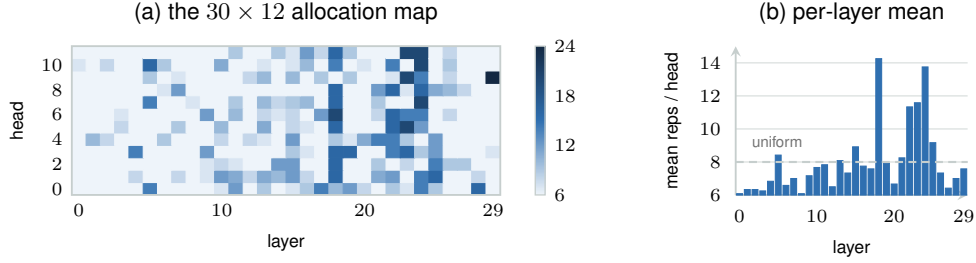
\begin{figure}[t]
    \centering
    \definecolor{obsblue}{HTML}{2F6FB0}
    \definecolor{obsrule}{HTML}{C6CECD}
    \definecolor{obsgrid}{HTML}{E2E6E5}
    \begin{minipage}[t]{0.60\linewidth}
        \centering
        \vspace{0pt}
        \begin{tikzpicture}
            \begin{axis}[
                width=7.25cm,height=3.55cm,
                title={(a) the $30\times12$ allocation map},
                title style={font=\sffamily\footnotesize,yshift=-1pt},
                enlargelimits=false,axis on top,
                xmin=-0.5,xmax=29.5,ymin=-0.5,ymax=11.5,
                xlabel={layer},ylabel={head},
                xtick={0,10,20,29},ytick={0,2,4,6,8,10},
                yticklabel style={font=\sffamily\scriptsize},xticklabel style={font=\sffamily\scriptsize},
                xlabel style={font=\sffamily\scriptsize},ylabel style={font=\sffamily\scriptsize},
                axis line style={draw=obsrule,line width=0.7pt},tick style={draw=none},
                colormap={cmblue}{rgb255=(238,244,250) rgb255=(47,111,176) rgb255=(19,40,66)},
                point meta min=6,point meta max=24,
                colorbar,colorbar style={width=0.13cm,font=\sffamily\scriptsize,ytick={6,12,18,24},
                    tick style={draw=none},axis line style={draw=obsrule,line width=0.5pt}},
            ]
            \addplot[matrix plot*,mesh/cols=30,point meta=explicit] table[x=layer,y=head,meta=v] {figures/assets/map_grid_T.dat};
            \end{axis}
        \end{tikzpicture}
    \end{minipage}\hfill
    \begin{minipage}[t]{0.39\linewidth}
        \centering
        \vspace{0pt}
        \begin{tikzpicture}
            \begin{axis}[
                width=4.65cm,height=3.55cm,
                title={(b) per-layer mean},
                title style={font=\sffamily\footnotesize,yshift=-1pt},
                xmin=-0.5,xmax=29.5,xtick={0,10,20,29},
                ymin=6,ymax=15,ytick={6,8,10,12,14},
                xlabel={layer},ylabel={mean reps / head},
                yticklabel style={font=\sffamily\scriptsize},xticklabel style={font=\sffamily\scriptsize},
                xlabel style={font=\sffamily\scriptsize},ylabel style={font=\sffamily\scriptsize},
                axis line style={draw=obsrule,line width=0.7pt},tick style={draw=none},axis lines=left,
                ymajorgrids=true,grid style={draw=obsgrid,line width=0.6pt},
            ]
            \addplot[ybar,bar width=2.2pt,fill=obsblue,draw=obsblue] table[x=layer,y=mean] {figures/assets/map_rowmean.dat};
            \draw[obsrule,densely dashed,line width=0.8pt] (axis cs:-0.5,8) -- (axis cs:29.5,8);
            \node[font=\sffamily\fontsize{6}{7}\selectfont,text=black!55,anchor=south west] at (axis cs:0.4,8.35) {uniform};
            \end{axis}
        \end{tikzpicture}
    \end{minipage}
    \vspace{-4mm}
    \caption{Calibrated allocation across layers and attention heads.
    Panel (a) shows the representative count for each layer and head, with darker colors indicating larger allocations.
    Panel (b) shows the mean representative count for each layer, with the dashed line indicating uniform allocation.
    The resulting allocation follows no simple ordering across layers or heads.}
    \label{fig:map}
\end{figure}

\section{Experiments}

\subsection{Setup}
We evaluate on VBench prompts \citep{huang2024vbench} that are separate from the $50$ calibration prompts.
We use $R{=}8$ and $B{=}2880$.
This uniform setting reduces the logical temporal representative capacity by $5.6\times$.
The current implementation preallocates the full KV buffer, so realized memory usage is unchanged (\cref{app:efficiency}).
We use one fixed seed for each prompt across all methods and compare every result with the corresponding Dense video.
We report mean LPIPS over the full clip, the first $60$ frames, and the last $12$ frames, together with PSNR and SSIM.
We compare allocation methods using our merge operator (\cref{tab:allocation}) and compare merge operators under a fixed uniform budget (\cref{tab:operators}).

\begin{table}[t]
\centering
\begin{tabular}{lccccc}
\toprule
Allocation & LPIPS & early & tail & PSNR & SSIM \\
\midrule
Uniform ($R{=}8$)      & 0.149 & 0.015 & 0.258 & 19.3 & 0.778 \\
Random                 & 0.156 & 0.022 & 0.261 & 19.1 & 0.769 \\
Head-only              & 0.146 & 0.014 & 0.255 & 19.4 & 0.781 \\
Layer-only             & 0.141 & 0.018 & 0.238 & 19.4 & 0.781 \\
Sensitivity            & 0.124 & 0.018 & 0.194 & 20.0 & 0.792 \\
Greedy                 & 0.118 & 0.015 & 0.191 & 20.1 & 0.797 \\
\midrule
\method (ours)         & \textbf{0.117} & 0.015 & \textbf{0.192} & \textbf{20.2} & \textbf{0.798} \\
\method-warm (ours)    & 0.131 & 0.015 & 0.211 & 19.8 & 0.789 \\
\bottomrule
\end{tabular}
\caption{Comparison of allocation strategies under the same KV budget and merge operator.
LPIPS measures perceptual error over the full clip, while early and tail report the first and last temporal windows.
Lower LPIPS and higher PSNR and SSIM indicate better quality.
\method\ achieves the best overall quality, with most of the improvement in later frames.}
\label{tab:allocation}
\end{table}

\subsection{Main results}

\begin{figure}[t]
\centering
\includegraphics[width=0.92\linewidth,trim={0 0 0 14},clip]{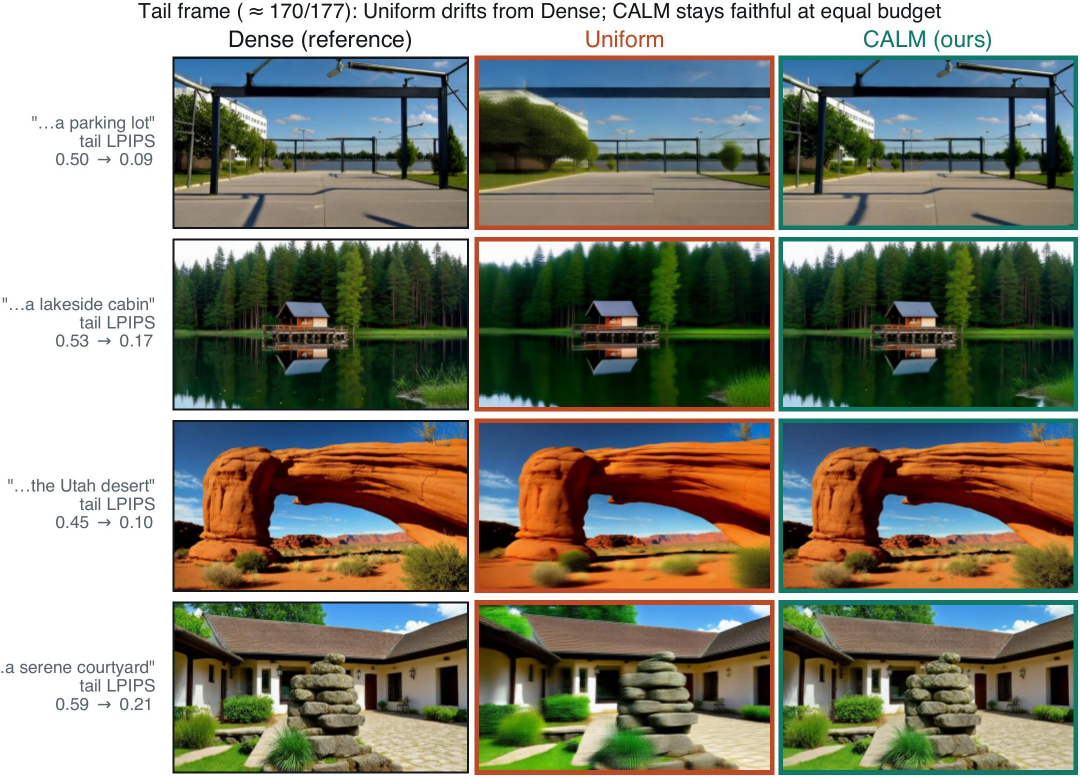}
\vspace{-2mm}
\caption{\textbf{Tail frame ($\approx\!170/177$): uniform drifts from Dense; CALM stays faithful at equal budget.} Late-rollout examples from VBench comparing Dense, uniform allocation, and CALM; the compressed methods share matched budgets and generation settings, and each row is annotated with its tail LPIPS.}
\label{fig:qual-tail}
\end{figure}

\paragraph{Calibrated allocation vs.\ uniform (same operator, same budget).}
We fix the merge operator to FPS and use the same total budget for all allocation methods.
At $R{=}8$ and $B{=}2880$, uniform merging obtains a full clip LPIPS of $0.149$ against the corresponding Dense videos on $400$ test prompts.
\method\ obtains an LPIPS of $0.117$ with a lower bound of $r_{\min}{=}6$ and improves the result on $356$ prompts (\cref{tab:allocation}).
Most of the improvement occurs in the later frames, as the qualitative comparison in \cref{fig:qual-tail} shows: uniform allocation visibly drifts in the late rollout while \method\ stays closer to Dense (further examples in \cref{app:qualitative}).
The LPIPS over the first $60$ frames is $0.015$ for both \method\ and uniform merging.
Greedy allocation obtains $0.118$, and the remaining allocation methods are compared in \cref{sec:ablation}.

\paragraph{VBench quality.}
\label{sec:vbench}
On six compression-relevant VBench dimensions, \method\ matches or improves on uniform merging and remains close to Dense (\cref{tab:vbench}). Each dimension uses its designated prompt subset, with matched prompts, seeds, and generation settings across methods; sample counts are reported in the table. Five additional dimensions appear in \cref{app:vbench-full}.
\begin{table}[t]
\centering
\begin{tabular}{lcccc}
\toprule
Dimension $\uparrow$ & $n$ & Dense & Uniform & \method \\
\midrule
Subject Consistency    & 72 & 0.948 & 0.928 & \textbf{0.941} \\
Background Consistency  & 85 & 0.955 & 0.947 & \textbf{0.952} \\
Temporal Flickering    & 75 & 0.990 & 0.989 & \textbf{0.990} \\
Motion Smoothness      & 72 & 0.984 & 0.983 & \textbf{0.984} \\
Imaging Quality        & 92 & 0.698 & 0.688 & \textbf{0.697} \\
Aesthetic Quality      & 92 & 0.628 & 0.613 & \textbf{0.622} \\
\midrule
VRAM save               & -- & $0\times$ & $5.6\times$ & $5.6\times$ \\
\bottomrule
\end{tabular}
\caption{VBench comparison on six dimensions relevant to cache compression.
$n$ is the number of matched samples, and higher scores indicate better quality.
VRAM save reports the reduction in logical cache capacity.
\method\ matches or outperforms uniform allocation on every dimension while remaining close to Dense.}
\label{tab:vbench}
\end{table}

\paragraph{Cross model generalization.}
\label{sec:generalization}
We apply \method\ to CausVid~\mbox{\citep{yin2024causvid}}, Causal Forcing~\mbox{\citep{zhu2026causalforcing}}, and a Self Gradient Forcing variant~\mbox{\citep{zhuang2026selfgradient}} on the Wan backbone.
We calibrate a new distortion table and DP allocation for each model.
\Cref{tab:crossmodel} compares uniform and calibrated allocation under the same budget within each model.
On CausVid, \method\ significantly reduces LPIPS at $R{=}2,3,4$, with an improvement of up to $16\%$ on all $30$ prompts at $R{=}2$, but provides no significant gain at $R{=}5,6$.
For Causal Forcing and Self Gradient Forcing, significant gains occur at both $R{=}4$ and $R{=}6$.
At $R{=}2$, the gain is not significant for Self Gradient Forcing; we do not read the apparent Causal Forcing result at $R{=}2$ as a genuine regression, because its uniform baseline is itself non-monotonic in $R$ (\cref{app:generalization}). The proxy error decreases for every model and budget, but the LPIPS improvement depends on the model and compression level (\cref{fig:mechanism,fig:budget}).
We therefore report the complete budget sweep for each model without averaging across models.
The same value of $R$ does not represent the same cache size or compression difficulty across models.
We also evaluate transfer of the Self Forcing allocation map to CausVid without recalibration in \cref{app:generalization}.

\begin{table}[t]
\centering
\small
\begin{tabular}{llccccc}
\toprule
Model & $R$ & Uniform $\downarrow$ & \method $\downarrow$ & rel. & $\Delta$ ($\method-$Uniform) & $n$ \\
\midrule
CausVid          & 2 & $0.135$ & $\mathbf{0.114}$ & $16\%$  & $\mathbf{-0.021}\ [-0.025,-0.018]$ & $30$ \\
                 & 3 & $0.107$ & $\mathbf{0.100}$ & $6\%$   & $\mathbf{-0.007}\ [-0.012,-0.003]$ & $30$ \\
                 & 4 & $0.096$ & $\mathbf{0.088}$ & $9\%$   & $\mathbf{-0.008}\ [-0.011,-0.005]$ & $30$ \\
                 & 5 & $0.083$ & $0.084$          & $-1\%$  & $+0.001\ [-0.001,+0.003]$          & $30$ \\
                 & 6 & $0.077$ & $0.078$          & $-1\%$  & $+0.001\ [-0.001,+0.003]$          & $30$ \\
\midrule
Causal Forcing   & 2 & $0.186$ & $0.223$          & $-20\%$ & $\mathbf{+0.037}\ [+0.012,+0.065]$ & $29$ \\
                 & 3 & $0.186$ & $0.189$          & $-2\%$  & $+0.003\ [-0.024,+0.027]$          & $29$ \\
                 & 4 & $0.212$ & $\mathbf{0.176}$ & $17\%$  & $\mathbf{-0.036}\ [-0.061,-0.011]$ & $29$ \\
                 & 6 & $0.196$ & $\mathbf{0.171}$ & $13\%$  & $\mathbf{-0.025}\ [-0.043,-0.010]$ & $29$ \\
\midrule
SGF  & 2 & $0.105$ & $0.107$          & $-2\%$  & $+0.002\ [-0.004,+0.007]$          & $29$ \\
                 & 3 & $0.087$ & $0.084$          & $3\%$   & $-0.003\ [-0.008,+0.003]$          & $29$ \\
                 & 4 & $0.091$ & $\mathbf{0.081}$ & $11\%$  & $\mathbf{-0.010}\ [-0.016,-0.004]$ & $29$ \\
                 & 6 & $0.085$ & $\mathbf{0.079}$ & $7\%$   & $\mathbf{-0.006}\ [-0.010,-0.002]$ & $29$ \\
\bottomrule
\end{tabular}
\caption{Comparison of uniform and calibrated allocation across autoregressive video generators after separate calibration for each model.
$R$ is the representative count, rel. is the relative LPIPS reduction, $\Delta$ is the paired LPIPS difference, and $n$ is the number of prompts.
Bold values indicate that the confidence interval for $\Delta$ excludes zero.
The effect of calibrated allocation depends on the model and compression level.}
\label{tab:crossmodel}
\end{table}



\subsection{Ablation and analysis}
\label{sec:ablation}

\paragraph{Allocation strategies.}
We fix the merge operator and compare six allocation strategies with uniform allocation in \cref{tab:allocation}.
The strategies are Random, Layer only, Head only, Sensitivity, Greedy, and the exact DP used by \method.
Layer only and Head only use one sensitivity axis, while Sensitivity uses one point from each distortion curve.
Greedy and \method\ use the complete distortion curves.
Greedy obtains an LPIPS of $0.118$, which is within $0.001$ of \method.
These results show that the measured joint sensitivity across layers and heads is more important than the choice of optimizer.

\begin{table}[t]
\centering
\begin{tabular}{lccccc}
\toprule
Operator (at $R{=}8$) & LPIPS & early & tail & PSNR & SSIM \\
\midrule
Random merging         & 0.339 & 0.082 & 0.534 & 13.2 & 0.582 \\
Average pooling        & 0.305 & 0.042 & 0.520 & 13.6 & 0.621 \\
AsymRnR-style (ported) & 0.190 & 0.019 & 0.333 & 17.1 & 0.726 \\
ToMe                   & 0.188 & 0.018 & 0.331 & 17.2 & 0.728 \\
DPC-kNN                & 0.187 & 0.017 & 0.336 & 17.3 & 0.731 \\
\midrule
FPS-based merging @ Uniform            & \textbf{0.149} & 0.015 & \textbf{0.258} & \textbf{19.3} & \textbf{0.778} \\
\midrule
\multicolumn{6}{l}{\emph{Allocation map transfers to ToMe without recalibration}}\\
ToMe @ uniform         & 0.188 & 0.018 & 0.331 & 17.2 & 0.728 \\
ToMe @ \method\ map    & 0.213 & 0.025 & 0.358 & 16.3 & 0.701 \\
\bottomrule
\end{tabular}
\caption{Comparison of temporal KV merge operators under the same uniform budget.
LPIPS measures perceptual error over the full clip, while early and tail report the first and last temporal windows.
The bottom rows evaluate direct transfer of the calibrated allocation map to ToMe.
FPS achieves the best quality among the uniform operators, while direct transfer to ToMe does not improve its results.}
\label{tab:operators}
\end{table}

\paragraph{Comparison with adapted merge implementations.}
\Cref{tab:operators} compares our method with adapted implementations of other merge operators under the same uniform budget.
These implementations use our pipeline rather than the original systems.
They differ from our method in grouping and key aggregation.
At the uniform budget, FPS obtains an LPIPS of $0.149$, ToMe obtains $0.188$ \citep{bolya2022tome}, DPC kNN obtains $0.187$, and the adapted AsymRnR method obtains $0.190$ \citep{sun2025asymrnr}.
Average pooling and random merging produce larger errors.
Full \method\ combines FPS with calibrated allocation and obtains $0.117$.
This comparison does not isolate the allocation gain because both the operator and allocation are different.

\paragraph{Proxy validation.}
\method\ selects the allocation map by minimizing the summed attention error $S(P){=}\sum_{l,h}D_{l,h}(P_{l,h})$.
Since the DP optimizes $S(P)$, the LPIPS results in \cref{tab:allocation} provide the main evidence for the allocation benefit; \cref{app:proxy} confirms that $S(P)$ nonetheless tracks video LPIPS across compression levels and across same-budget allocation maps.

\paragraph{Floors, temporal behavior, and budget sensitivity.}
\label{sec:floor-analysis}
The lower bound $r_{\min}$ specifies the minimum number of temporal representatives per layer--head. Increasing it restricts allocation flexibility without increasing the total budget. We compare $r_{\min}{=}2$ and $r_{\min}{=}6$ on the same held-out subset, using the same merge operator and calibration table and re-solving the DP at $B{=}2880$. With $r_{\min}{=}2$, \method\ improves full-clip LPIPS over uniform allocation ($0.058$ vs.\ $0.066$) but increases first-$60$-frame LPIPS ($0.020$ vs.\ $0.002$). Raising the floor to $6$ substantially reduces this early degradation ($0.003$) and further lowers full-clip LPIPS to $0.043$ (\cref{fig:floor}). The budget sweep in \cref{app:budget} examines how allocation gains vary with compression strength.

\paragraph{The calibrated map is irregular.}
The calibrated map is not monotonic across layer or head indices (\cref{fig:map}).
The remaining budget follows no simple ordering across layers or heads.
This irregular map further demonstrates the need for our method because an optimal budget allocation is complex and difficult to explain with simple heuristics.
It therefore requires direct calibration of the entire layer and head map.


\section{Conclusion}
We introduce \method, which calibrates layer--head sensitivity using attention-output error and solves a fixed-budget allocation offline. The frozen allocation improves fidelity over uniform merging on unseen prompts without per-video search. Cross-model experiments show that its benefits depend on the generator and compression level. The compression is logical: our implementation keeps $B$ representatives but pre-allocates the full KV buffer, so the memory ratio is a bound rather than a realized latency or memory win (\cref{tab:efficiency}).

\clearpage
\subsection*{AI use statement}
Generative AI assisted with experimental coding, idea exploration, and edits to the writing. 
The tools used included GPT-Codex and Claude. We checked the resulting code and text for correctness, accuracy, and suitability. The authors are fully responsible for the final manuscript, its claims, and all accompanying artifacts, including material developed with AI assistance.

\subsection*{Reproducibility statement}
Our experiments are reproducible, and we will release the code upon acceptance.

\bibliography{iclr2027_conference}

\begin{thebibliography}{33}
\providecommand{\natexlab}[1]{#1}
\providecommand{\url}[1]{\texttt{#1}}
\expandafter\ifx\csname urlstyle\endcsname\relax
  \providecommand{\doi}[1]{doi: #1}\else
  \providecommand{\doi}{doi: \begingroup \urlstyle{rm}\Url}\fi

\bibitem[Bolya \& Hoffman(2023)Bolya and Hoffman]{bolya2023token}
Daniel Bolya and Judy Hoffman.
\newblock Token merging for fast stable diffusion.
\newblock In \emph{Proceedings of the IEEE/CVF Conference on Computer Vision
  and Pattern Recognition (CVPR) Workshops}, 2023.

\bibitem[Bolya et~al.(2023)Bolya, Fu, Dai, Zhang, Feichtenhofer, and
  Hoffman]{bolya2022tome}
Daniel Bolya, Cheng-Yang Fu, Xiaoliang Dai, Peizhao Zhang, Christoph
  Feichtenhofer, and Judy Hoffman.
\newblock Token merging: Your {ViT} but faster.
\newblock In \emph{International Conference on Learning Representations}, 2023.

\bibitem[Brooks et~al.(2024)Brooks, Peebles, Holmes, DePue, Guo, Jing, Schnurr,
  Taylor, Luhman, Luhman, Ng, Wang, and Ramesh]{brooks2024sora}
Tim Brooks, Bill Peebles, Connor Holmes, Will DePue, Yufei Guo, Li~Jing, David
  Schnurr, Joe Taylor, Troy Luhman, Eric Luhman, Clarence Ng, Ricky Wang, and
  Aditya Ramesh.
\newblock Video generation models as world simulators.
\newblock
  \url{https://openai.com/index/video-generation-models-as-world-simulators/},
  2024.

\bibitem[Chen et~al.(2024)Chen, Mart{\'i}~Mons{\'o}, Du, Simchowitz, Tedrake,
  and Sitzmann]{chen2024diffusionforcing}
Boyuan Chen, Diego Mart{\'i}~Mons{\'o}, Yilun Du, Max Simchowitz, Russ Tedrake,
  and Vincent Sitzmann.
\newblock Diffusion forcing: Next-token prediction meets full-sequence
  diffusion.
\newblock In \emph{Advances in Neural Information Processing Systems}, 2024.

\bibitem[Chen et~al.(2023)Chen, Shao, Xu, Lin, Zhang, Chao, Ji, Qiao, and
  Luo]{chen2023diffrate}
Mengzhao Chen, Wenqi Shao, Peng Xu, Mingbao Lin, Kaipeng Zhang, Fei Chao,
  Rongrong Ji, Yu~Qiao, and Ping Luo.
\newblock {DiffRate}: Differentiable compression rate for efficient vision
  transformers.
\newblock In \emph{Proceedings of the IEEE/CVF International Conference on
  Computer Vision (ICCV)}, 2023.

\bibitem[Choi et~al.(2024)Choi, Lee, Chu, Choi, and Kim]{choi2024vidtldr}
Joonmyung Choi, Sanghyeok Lee, Jaewon Chu, Minhyuk Choi, and Hyunwoo~J. Kim.
\newblock {vid-TLDR}: Training free token merging for light-weight video
  transformer.
\newblock In \emph{Proceedings of the IEEE/CVF Conference on Computer Vision
  and Pattern Recognition (CVPR)}, 2024.

\bibitem[Choudhury et~al.(2024)Choudhury, Zhu, Liu, Niinuma, Kitani, and
  Jeni]{choudhury2024dont}
Rohan Choudhury, Guanglei Zhu, Sihan Liu, Koichiro Niinuma, Kris~M. Kitani, and
  L{\'a}szl{\'o} Jeni.
\newblock Don't look twice: Faster video transformers with run-length
  tokenization.
\newblock In \emph{Advances in Neural Information Processing Systems
  (NeurIPS)}, 2024.

\bibitem[Hu et~al.(2026)Hu, Wang, He, Gong, Yi, Zhang, Bai, Chen, Zhang, Li,
  and Yuan]{hu2026centroidkv}
Jie Hu, Shengnan Wang, Yutong He, Ping Gong, Jiawei Yi, Juncheng Zhang, Youhui
  Bai, Renhai Chen, Gong Zhang, Cheng Li, and Kun Yuan.
\newblock {CentroidKV}: Efficient long-context {LLM} inference via {KV} cache
  clustering.
\newblock \emph{Transactions on Machine Learning Research}, 2026.

\bibitem[Huang et~al.(2025)Huang, Li, He, Zhou, and
  Shechtman]{huang2025selfforcing}
Xun Huang, Zhengqi Li, Guande He, Mingyuan Zhou, and Eli Shechtman.
\newblock Self forcing: Bridging the train-test gap in autoregressive video
  diffusion.
\newblock In \emph{Advances in Neural Information Processing Systems}, 2025.

\bibitem[Huang et~al.(2024)Huang, He, Yu, Zhang, Si, Jiang, Zhang, Wu, Jin,
  Chanpaisit, Wang, Chen, Wang, Lin, Qiao, and Liu]{huang2024vbench}
Ziqi Huang, Yinan He, Jiashuo Yu, Fan Zhang, Chenyang Si, Yuming Jiang, Yuanhan
  Zhang, Tianxing Wu, Qingyang Jin, Nattapol Chanpaisit, Yaohui Wang, Xinyuan
  Chen, Limin Wang, Dahua Lin, Yu~Qiao, and Ziwei Liu.
\newblock {VBench}: Comprehensive benchmark suite for video generative models.
\newblock In \emph{Proceedings of the IEEE/CVF Conference on Computer Vision
  and Pattern Recognition (CVPR)}, 2024.

\bibitem[Ji et~al.(2026)Ji, Zhong, Zhang, Yang, Jin, Qin, Luo, Mao, Liu, and
  Li]{ji2026forcingkv}
Yicheng Ji, Zhizhou Zhong, Jun Zhang, Qin Yang, XiTai Jin, Ying Qin, Wenhan
  Luo, Shuiyang Mao, Wei Liu, and Huan Li.
\newblock Forcing-kv: Hybrid kv cache compression for efficient autoregressive
  video diffusion models.
\newblock \emph{arXiv preprint arXiv:2605.09681}, 2026.

\bibitem[Li et~al.(2024{\natexlab{a}})Li, Ma, Yang, and Yang]{li2024vidtome}
Xirui Li, Chao Ma, Xiaokang Yang, and Ming-Hsuan Yang.
\newblock {VidToMe}: Video token merging for zero-shot video editing.
\newblock In \emph{IEEE/CVF Conference on Computer Vision and Pattern
  Recognition}, 2024{\natexlab{a}}.

\bibitem[Li et~al.(2024{\natexlab{b}})Li, Huang, Yang, Venkitesh, Locatelli,
  Ye, Cai, Lewis, and Chen]{li2024snapkv}
Yuhong Li, Yingbing Huang, Bowen Yang, Bharat Venkitesh, Acyr Locatelli,
  Hanchen Ye, Tianle Cai, Patrick Lewis, and Deming Chen.
\newblock {SnapKV}: {LLM} knows what you are looking for before generation.
\newblock In \emph{Advances in Neural Information Processing Systems},
  2024{\natexlab{b}}.

\bibitem[Liang et~al.(2022)Liang, Ge, Tong, Song, Wang, and Xie]{liang2022not}
Youwei Liang, Chongjian Ge, Zhan Tong, Yibing Song, Jue Wang, and Pengtao Xie.
\newblock Not all patches are what you need: Expediting vision transformers via
  token reorganizations.
\newblock In \emph{International Conference on Learning Representations
  (ICLR)}, 2022.

\bibitem[Liu et~al.(2024)Liu, Liu, Pan, He, Haffari, and
  Zhuang]{liu2024minicache}
Akide Liu, Jing Liu, Zizheng Pan, Yefei He, Gholamreza Haffari, and Bohan
  Zhuang.
\newblock {MiniCache}: {KV} cache compression in depth dimension for large
  language models.
\newblock In \emph{Advances in Neural Information Processing Systems
  (NeurIPS)}, 2024.

\bibitem[Lu et~al.(2025)Lu, Zheng, Xia, and Wang]{lu2025toma}
Wenbo Lu, Shaoyi Zheng, Yuxuan Xia, and Shengjie Wang.
\newblock {ToMA}: Token merge with attention for diffusion models.
\newblock In \emph{Proceedings of the 42nd International Conference on Machine
  Learning (ICML)}, 2025.

\bibitem[Sun et~al.(2025)Sun, Tu, Liao, Jin, and Tao]{sun2025asymrnr}
Wenhao Sun, Rong-Cheng Tu, Jingyi Liao, Zhao Jin, and Dacheng Tao.
\newblock {AsymRnR}: Video diffusion transformers acceleration with asymmetric
  reduction and restoration.
\newblock In \emph{Proceedings of the 42nd International Conference on Machine
  Learning (ICML)}, 2025.

\bibitem[Voita et~al.(2019)Voita, Talbot, Moiseev, Sennrich, and
  Titov]{voita2019analyzing}
Elena Voita, David Talbot, Fedor Moiseev, Rico Sennrich, and Ivan Titov.
\newblock Analyzing multi-head self-attention: Specialized heads do the heavy
  lifting, the rest can be pruned.
\newblock \emph{arXiv preprint arXiv:1905.09418}, 2019.

\bibitem[Wan et~al.(2025{\natexlab{a}})Wan, Wang, Ai, et~al.]{wan2025}
Team Wan, Ang Wang, Baole Ai, et~al.
\newblock Wan: Open and advanced large-scale video generative models.
\newblock \emph{arXiv preprint arXiv:2503.20314}, 2025{\natexlab{a}}.

\bibitem[Wan et~al.(2025{\natexlab{b}})Wan, Wu, Zhang, Xin, Tao, Zhu, Wang,
  Luo, Xiong, Wang, and Zhang]{wan2025d2o}
Zhongwei Wan, Xinjian Wu, Yu~Zhang, Yi~Xin, Chaofan Tao, Zhihong Zhu, Xin Wang,
  Siqi Luo, Jing Xiong, Longyue Wang, and Mi~Zhang.
\newblock {D2O}: Dynamic discriminative operations for efficient long-context
  inference of large language models.
\newblock In \emph{International Conference on Learning Representations
  (ICLR)}, 2025{\natexlab{b}}.

\bibitem[Wang et~al.(2024)Wang, Jin, Yu, and Zhang]{wang2024model}
Zheng Wang, Boxiao Jin, Zhongzhi Yu, and Minjia Zhang.
\newblock Model tells you where to merge: Adaptive {KV} cache merging for
  {LLMs} on long-context tasks, 2024.

\bibitem[Wu et~al.(2025)Wu, Xu, Le, and Samaras]{wu2025importance}
Haoyu Wu, Jingyi Xu, Hieu Le, and Dimitris Samaras.
\newblock Importance-based token merging for efficient image and video
  generation.
\newblock In \emph{IEEE/CVF International Conference on Computer Vision}, 2025.

\bibitem[Xiao et~al.(2024)Xiao, Tian, Chen, Han, and
  Lewis]{xiao2024streamingllm}
Guangxuan Xiao, Yuandong Tian, Beidi Chen, Song Han, and Mike Lewis.
\newblock Efficient streaming language models with attention sinks.
\newblock In \emph{International Conference on Learning Representations}, 2024.

\bibitem[Xu et~al.(2026)Xu, Du, Liu, Yang, Jiang, Yan, Saha, Pumarola, Wang,
  and Li]{xu2026sparseforcing}
Boxun Xu, Yuming Du, Zichang Liu, Siyu Yang, Ziyang Jiang, Siqi Yan, Rajasi
  Saha, Albert Pumarola, Wenchen Wang, and Peng Li.
\newblock Sparse forcing: Native trainable sparse attention for real-time
  autoregressive diffusion video generation.
\newblock \emph{arXiv preprint arXiv:2604.21221}, 2026.

\bibitem[Yang et~al.(2025)Yang, Teng, Zheng, Ding, Huang, Xu, Yang, Hong,
  Zhang, Feng, Yin, Zhang, Wang, Cheng, Xu, Gu, Dong, and
  Tang]{yang2025cogvideox}
Zhuoyi Yang, Jiayan Teng, Wendi Zheng, Ming Ding, Shiyu Huang, Jiazheng Xu,
  Yuanming Yang, Wenyi Hong, Xiaohan Zhang, Guanyu Feng, Da~Yin, Yuxuan Zhang,
  Weihan Wang, Yean Cheng, Bin Xu, Xiaotao Gu, Yuxiao Dong, and Jie Tang.
\newblock {CogVideoX}: Text-to-video diffusion models with an expert
  transformer.
\newblock In \emph{International Conference on Learning Representations}, 2025.

\bibitem[Yin et~al.(2024)Yin, Gharbi, Zhang, Shechtman, Durand, Freeman, and
  Park]{yin2024dmd}
Tianwei Yin, Micha{\"e}l Gharbi, Richard Zhang, Eli Shechtman, Fr{\'e}do
  Durand, William~T Freeman, and Taesung Park.
\newblock One-step diffusion with distribution matching distillation.
\newblock In \emph{Proceedings of the IEEE/CVF Conference on Computer Vision
  and Pattern Recognition}, 2024.

\bibitem[Yin et~al.(2025)Yin, Zhang, Zhang, Freeman, Durand, Shechtman, and
  Huang]{yin2024causvid}
Tianwei Yin, Qiang Zhang, Richard Zhang, William~T Freeman, Fr{\'e}do Durand,
  Eli Shechtman, and Xun Huang.
\newblock From slow bidirectional to fast autoregressive video diffusion
  models.
\newblock In \emph{IEEE/CVF Conference on Computer Vision and Pattern
  Recognition}, 2025.

\bibitem[Yuan et~al.(2025)Yuan, He, Bai, Leng, and Jiang]{yuan2025weightedkv}
Jian Yuan, Ziwei He, Haoli Bai, Jingwen Leng, and Bo~Jiang.
\newblock {WeightedKV}: Attention scores weighted key-value cache merging for
  large language models.
\newblock In \emph{IEEE International Conference on Acoustics, Speech and
  Signal Processing (ICASSP)}, 2025.

\bibitem[Zeng et~al.(2022)Zeng, Jin, Liu, Qian, Luo, Ouyang, and
  Wang]{zeng2022not}
Wang Zeng, Sheng Jin, Wentao Liu, Chen Qian, Ping Luo, Wanli Ouyang, and
  Xiaogang Wang.
\newblock Not all tokens are equal: Human-centric visual analysis via token
  clustering transformer.
\newblock In \emph{Proceedings of the IEEE/CVF Conference on Computer Vision
  and Pattern Recognition (CVPR)}, 2022.

\bibitem[Zhang et~al.(2024)Zhang, Du, Luo, Zhong, Zhang, Liu, and
  Ji]{zhang2024cam}
Yuxin Zhang, Yuxuan Du, Gen Luo, Yunshan Zhong, Zhenyu Zhang, Shiwei Liu, and
  Rongrong Ji.
\newblock {CaM}: Cache merging for memory-efficient {LLMs} inference.
\newblock In \emph{Proceedings of the 41st International Conference on Machine
  Learning (ICML)}, 2024.

\bibitem[Zhang et~al.(2023)Zhang, Sheng, Zhou, Chen, Zheng, Cai, Song, Tian,
  R{\'e}, Barrett, Wang, and Chen]{zhang2023h2o}
Zhenyu Zhang, Ying Sheng, Tianyi Zhou, Tianlong Chen, Lianmin Zheng, Ruisi Cai,
  Zhao Song, Yuandong Tian, Christopher R{\'e}, Clark Barrett, Zhangyang Wang,
  and Beidi Chen.
\newblock {H2O}: Heavy-hitter oracle for efficient generative inference of
  large language models.
\newblock In \emph{Advances in Neural Information Processing Systems}, 2023.

\bibitem[Zhu et~al.(2026)Zhu, Zhao, He, Su, Li, and Zhu]{zhu2026causalforcing}
Hongzhou Zhu, Min Zhao, Guande He, Hang Su, Chongxuan Li, and Jun Zhu.
\newblock Causal forcing: Autoregressive diffusion distillation done right for
  high-quality real-time interactive video generation.
\newblock In \emph{International Conference on Machine Learning}, 2026.

\bibitem[Zhuang et~al.(2026)Zhuang, Zhang, Bian, Li, Luo, Liu, Jin, Zhang, He,
  Zhang, Li, Huang, Xue, and Duan]{zhuang2026selfgradient}
Junhao Zhuang, Shiyi Zhang, Yuxuan Bian, Yaowei Li, Yawen Luo, Yijun Liu,
  Weiyang Jin, Songchun Zhang, Xianglong He, Xuying Zhang, Haoran Li, Haoyang
  Huang, Zeyue Xue, and Nan Duan.
\newblock Self gradient forcing: Native long video extrapolation.
\newblock \emph{arXiv preprint arXiv:2607.20368}, 2026.

\end{thebibliography}
\bibliographystyle{iclr2027_conference}

\appendix

\clearpage
\section{Qualitative results}
\label{app:qualitative}
\begin{figure}[!ht]
    \centering
    \includegraphics[width=0.75\linewidth]{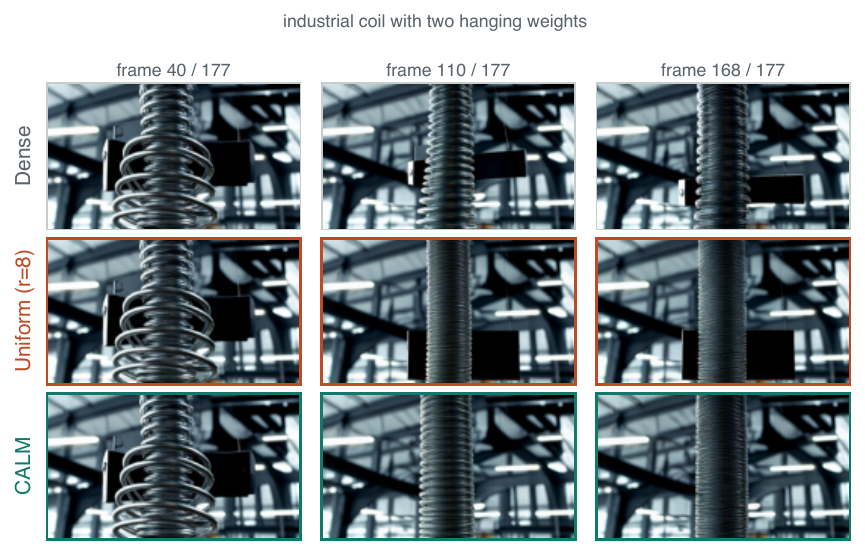}\\[5pt]
    \includegraphics[width=0.75\linewidth]{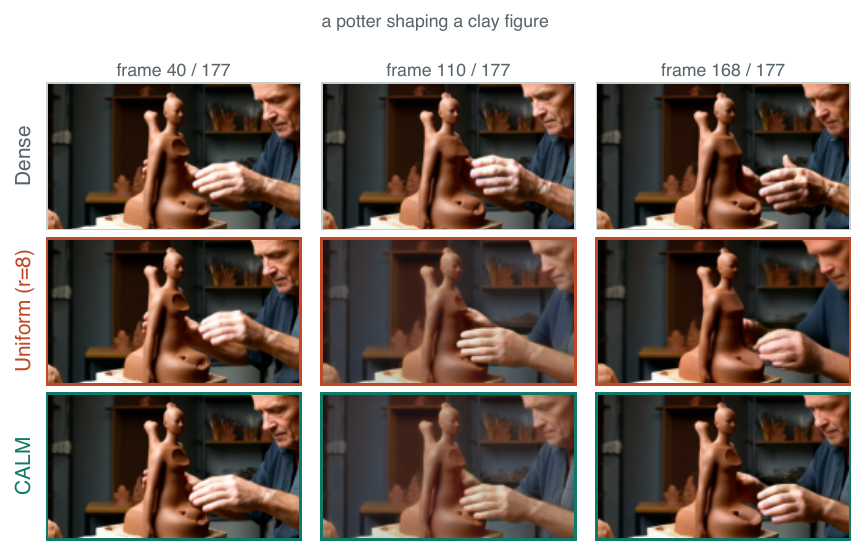}
    \caption{Where the difference emerges: two held-out prompts (top and bottom blocks), each shown at early, middle, and late rollout frames (columns).
    Within a block the rows are the Dense reference, uniform allocation, and \method; all three share the prompt and seed, and the two compressed rows share the representative budget.
    The arms are indistinguishable early and separate in the later frames, where \method\ stays closer to Dense.}
    \label{fig:qual}
\end{figure}

We collect additional video comparisons in this section, extending the qualitative result shown in the main text (\cref{fig:qual-tail}). In every panel the three arms are the uncompressed Dense reference, uniform allocation, and \method. The two compressed arms share the prompt, seed, generation settings, and \emph{representative budget}, so any difference between them reflects how that budget is distributed; Dense is the no-compression reference and is not budget-matched. Frames are sampled at the rollout positions marked in each figure, and because the differences are temporal and accumulate over the rollout, still frames understate the tail-LPIPS gap that the numbers report. The rows are illustrative examples from the held-out and VBench-400 test prompts, not a uniform random sample.

\begin{figure}[htbp]
\centering
\includegraphics[width=0.92\linewidth]{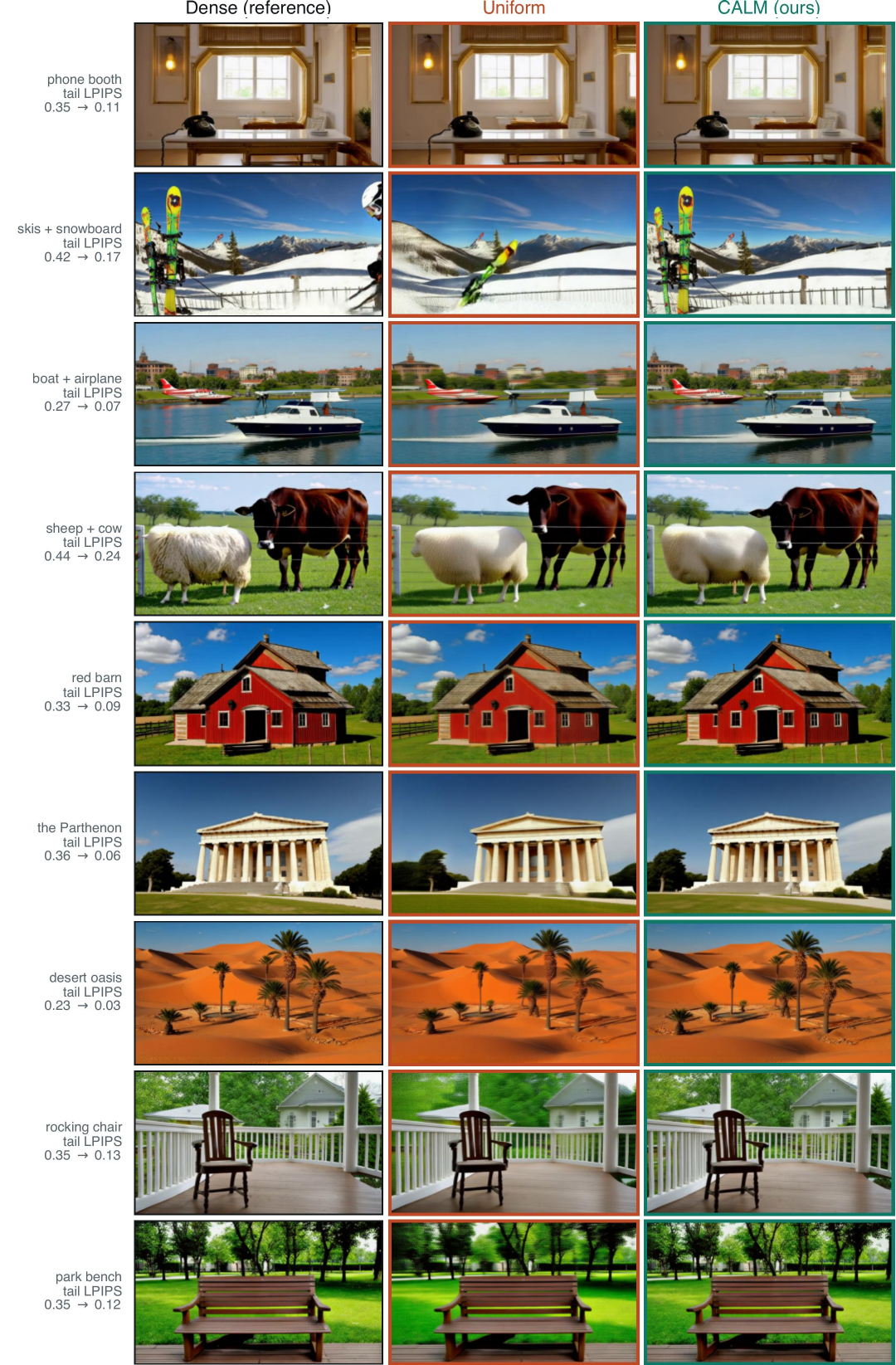}
\caption{Additional prompts from the VBench-400 test set at a late rollout frame, spanning objects, animals, architecture, and landscapes.
Columns are the Dense reference, uniform allocation, and \method; the two compressed columns share the representative budget and seed, and Dense is the uncompressed reference.
Rows are annotated with the tail-LPIPS gap between uniform and \method\ (per row) and span clear improvements to near-ties.}
\label{fig:qual-gallery2}
\end{figure}

\begin{figure}[htbp]
    \centering
    \includegraphics[width=0.92\linewidth]{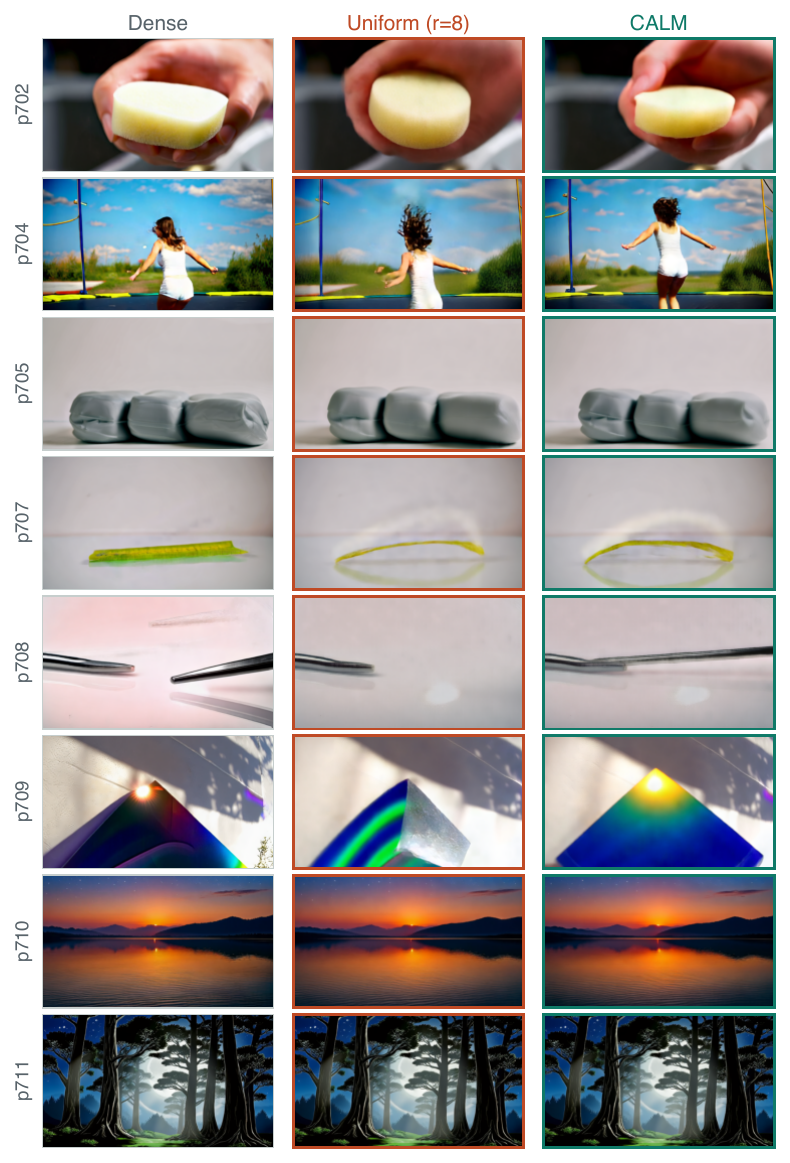}
    \caption{Eight further held-out prompts (disjoint from the calibration set) at a late rollout frame, covering close-up objects, human motion, landscapes, and stylized scenes.
    Columns are the Dense reference, uniform allocation, and \method; the two compressed columns share the representative budget and seed, and Dense is the uncompressed reference.
    These are illustrative cases where uniform allocation degrades visibly; on them \method\ stays closer to Dense.}
    \label{fig:gallery}
\end{figure}

\Cref{fig:qual-tail} shows late-rollout frames on VBench-400 test prompts (\cref{tab:allocation}) annotated with each row's tail LPIPS; \cref{fig:qual} tracks two held-out prompts from early to late frames to show when the gap emerges; \cref{fig:qual-gallery2} extends the test-set comparison to nine further prompts; and \cref{fig:gallery} adds eight further held-out prompts (disjoint from the calibration set) at a late frame.

\clearpage
\section{Experimental protocol}
\label{app:protocol}

\paragraph{Models and generation.}
Our main experiments use Self-Forcing on the Wan2.1-T2V-1.3B backbone; the cross-model study (\cref{app:generalization}) adds CausVid, Causal Forcing, and a Self-Gradient-Forcing variant. Within any comparison all arms share the prompt, seed, and sampler settings and use the same mask-aware SDPA attention backend, and each comparison varies a single factor---the allocation map, or the merge operator (which also changes grouping and key aggregation)---while holding the others fixed. Dense denotes the uncompressed reference.

\paragraph{Prompt sets.}
We use $50$ prompts for calibration (\cref{app:calibration}) and evaluate the main allocation table (\cref{tab:allocation}) on a $400$-prompt VBench subset disjoint from the calibration set. The proxy pre-registration (\cref{app:proxy}) uses a separate set of $20$ prompts, split into $12$ for choosing the error definition and $8$ for confirmation. The qualitative galleries (\cref{app:qualitative}) use held-out prompts as noted per figure. Cross-model evaluations use each model's own held-out prompts ($30$ for CausVid, $29$ for Causal Forcing and SGF).

\paragraph{Metrics.}
Unless stated otherwise, quality is per-frame LPIPS against the Dense reference over the full clip, with the first- and last-window (early/tail) LPIPS reported separately because the error is end-loaded; we also report PSNR and SSIM. VBench dimensions are scored with the official VBench scorers on their respective prompt subsets (\cref{app:vbench-full}).

\paragraph{Statistics.}
Cross-arm comparisons are paired over shared prompts. We report the mean paired difference against uniform merging, its $95\%$ confidence interval from a paired bootstrap over prompts, and the number of prompts on which the calibrated map wins; a difference is called significant when its confidence interval excludes zero.

\section{Calibration and allocation details}
\label{app:calibration}

We give the exact procedure that turns a set of calibration prompts and a frozen model into the
$30\times12$ allocation map; \cref{alg:calibrate,alg:allocate} give the pseudocode. The calibration fixes the
prompts, seeds, and implementation settings and sweeps a fixed grid of representative counts.

\paragraph{Probing one representative count.}
For a single representative count $r$, we generate a calibration prompt \emph{densely} (merging off) and, alongside
the dense cache, maintain a shadow KV cache: a recursively merged copy of the dense history,
built with our operator (FPS on the keys, count-weighted mean value, $\log$-count bias). After each
chunk's clean-context pass, that chunk's clean K/V join the shadow KV cache with unit counts and it is merged
again to $r$, so counts carry across chunks exactly as in merged generation. At each attention call---$15$ blocks
$\times$ $5$ passes (four denoising steps plus the clean-context pass)---we re-run the block's own
query against the shadow KV cache plus the current chunk's dense K/V and compare the resulting attention output $O^{(r)}_{l,h}$ against the dense output
$O^{\mathrm{full}}_{l,h}$. Because the shadow KV cache holds all $L\times H$ heads, one dense generation
measures the merge error of all $360$ heads at that $r$ simultaneously. No budget enters here:
each $(l,h,r)$ is probed independently; the budget $B$ only enters the allocation step.

\paragraph{Filling one prompt's table.}
The error of a head is its \emph{relative} MSE, \cref{eq:relmse}: the squared attention-output error summed
over batch, tokens, channels, and the whole rollout (every block and pass), divided by the squared dense output
summed the same way. The implementation clamps the denominator to at least $10^{-30}$ for numerical stability; this safeguard is omitted from \cref{eq:relmse}.
Sweeping $r$ over the grid $\mathcal{R}$ ($17$ values, $2$ to $41$, denser near the
uniform $r{=}8$) fills one prompt's table $E[l,h,r]$ at a cost of $|\mathcal{R}|$ dense generations per prompt.

\paragraph{Integrating prompts and solving.}
We average the per-prompt relative MSE over the calibration set, $D^{\mathrm{raw}}[l,h,r]=\tfrac{1}{|\mathcal{C}|}\sum_{x}
E_x[l,h,r]$, and take a running minimum along the full measured grid $\mathcal{R}$ so the \emph{table used for optimization},
$D$, is monotone in $r$
(a smoothing of the calibration table, not a guarantee about the error of any individual generation). Averaging
ratios rather than pooling numerators and denominators weights every prompt equally. We use $50$ calibration
prompts; the resulting DP allocation reproduces our deployed map, and the sensitivity ranking is stable across
prompts (\cref{fig:obs-consistency}). The dynamic program of \cref{alg:allocate} then keeps only the columns of $D$ in
$\mathcal{R}_{\mathrm{search}}=\{r\in\mathcal{R}:r_{\min}\le r\le r_{\max}\}$, after the running minimum, and allocates
the budget $B$ over them; the frozen map is applied to all unseen prompts.

\begin{algorithm}[t]
\caption{Budget Allocation}
\label{alg:allocate}
\definecolor{codecomment}{rgb}{0.10,0.55,0.10}
\definecolor{codekeyword}{rgb}{0.70,0.12,0.45}
\lstset{
    backgroundcolor=\color{white},
    basicstyle=\fontsize{7pt}{7.5pt}\ttfamily\selectfont,
    columns=fullflexible,
    breaklines=true,
    commentstyle=\color{codecomment},
    keywordstyle=\color{codekeyword},
    showstringspaces=false
}

\begin{lstlisting}[language=Python]
# Input: processed distortion table D with shape [L, H, |R|] (Alg. 2);
#        measured grid R; bounds r_min, r_max; budget B
# Output: integer allocation map P with shape [L, H], sum(P) = B

# Step 1: keep admissible counts, flatten layer-head pairs
keep = [j for j, r in enumerate(R) if r_min <= r <= r_max]
R = [R[j] for j in keep]  # R_search
L, H = D.shape[:2]
N = L * H
D = reshape(D[:, :, keep], (N, len(R)))
# Step 2: fill the DP table
DP = full((N + 1, B + 1), infinity)
DP[0, 0] = 0
choice = zeros((N + 1, B + 1), dtype=int)
for i in range(1, N + 1):
    for t in range(min(R), B + 1):
        DP[i, t], choice[i, t] = min(
            (DP[i-1, t-r] + D[i-1, j], r)
            for j, r in enumerate(R) if r <= t)
# Step 3: recover the allocation map
P = zeros(N, dtype=int)
t = B
for i in reversed(range(1, N + 1)):
    P[i-1] = choice[i, t]
    t -= P[i-1]

return reshape(P, (L, H))
\end{lstlisting}
\end{algorithm}

\section{Proxy evaluation}
\label{app:proxy}

Re-generating a full video to score every candidate merge is the bottleneck of any allocation search: across many layers, heads, and admissible representative counts, the space of same-budget allocation maps is far too large to explore one video at a time.
If instead a cheap, per-(layer, head) attention-output error ranks merge configurations in the same order that full-video LPIPS does, the search can be run entirely offline, evaluating each candidate from cached attention statistics rather than a generated video.
We use the per-(layer, head) relative mean-squared error between the merged and dense attention outputs, \cref{eq:relmse}, so that a single scalar per (layer, head) summarizes a candidate.

We fixed the analysis before inspecting any correlation.
The decision rule was pre-registered as a per-prompt Spearman correlation of at least $0.8$ between attention error and LPIPS.
We partitioned the $20$ prompts into $12$ ``choosing'' prompts, used only to select the error definition, and $8$ held-out ``confirmation'' prompts, reserved to test that choice.
We also pre-registered $17$ admissible representative counts, spaced more densely near the uniform setting $r{=}8$.

Two analyses probe the proxy, and they answer different questions on different samples and units; we report them separately rather than pooling them into one summary.
The first, within-prompt analysis correlates, for each of the $20$ pre-registered prompts, the attention error against video LPIPS across the $17$ \emph{uniform} representative counts; because the total budget varies with the count, it tests only whether the proxy tracks LPIPS as overall compression strength changes, not whether it ranks allocations at a fixed budget. The median per-prompt Spearman correlation is $1.00$, the minimum is $0.96$, and all $20$ prompts clear the pre-registered $0.8$ threshold. Separately, \cref{fig:proxy} pools $680$ prompt--count pairs into a single log--log correlation (Spearman $\rho{=}0.90$, Pearson $r{=}0.94$); this pooled value is lower than the per-prompt median; the pooled and per-prompt statistics use different scopes and are not directly comparable.
The second, same-budget analysis is the one that bears directly on the allocation search: it fixes the total budget at $B{=}2880$ and asks whether the offline score orders \emph{different allocations of the same budget} the way measured LPIPS does. Across seven static, equal-budget allocations---uniform, the DP map, and five variants, all at $B{=}2880$ with the same merge operator and bounds, evaluated on the same $102$ common prompts---the summed score $S(P){=}\sum_{l,h}D_{l,h}(P_{l,h})$ correlates strongly with mean video LPIPS (Spearman $\rho{=}0.96$, Pearson $r{=}0.98$; \cref{tab:proxy-maps}). The single rank inversion is the near-tie between the DP and greedy maps, whose scores differ by about $10^{-4}$. The $102$ prompts are those on which all seven maps completed generation (the intersection of the per-map evaluation sets). The distortion table and its running-minimum processing, the bounds, and all seven candidate maps were fixed by the main calibration before these generations were scored, so no hyperparameter was selected on this comparison; it is a post-hoc validation of the proxy rather than a tuning step.

\begin{table}[H]
\centering
\small
\begin{tabular}{lcc}
\toprule
Allocation & $S(P)$ & mean LPIPS $\downarrow$ \\
\midrule
\method\ (DP)       & $4.4550$ & $0.1684$ \\
Greedy              & $4.4551$ & $0.1683$ \\
Sensitivity         & $4.8466$ & $0.1789$ \\
Layer-only          & $5.6131$ & $0.1999$ \\
Head-only           & $6.7252$ & $0.2095$ \\
Uniform ($R{=}8$)   & $6.7315$ & $0.2143$ \\
Random              & $7.5905$ & $0.2223$ \\
\bottomrule
\end{tabular}
\caption{Offline calibration score $S(P)=\sum_{l,h}D_{l,h}(P_{l,h})$ versus end-to-end video LPIPS for seven static, equal-budget allocations of the same budget ($B{=}2880$), evaluated with the same merge operator and bounds on the same $102$ common prompts.
The score orders the allocations in nearly the same order as measured LPIPS (Spearman $\rho{=}0.96$, Pearson $r{=}0.98$).
The single inversion is the near-tie between the \method\ (DP) and Greedy maps, whose scores differ by about $10^{-4}$; extra decimals are shown to expose it.}
\label{tab:proxy-maps}
\end{table}

On the within-prompt uniform sweep, all $18$ candidate error definitions --- relative MSE, $L_1$, or cosine distance, under three poolings, computed per head or pooled across heads --- yield the same verdict, so that conclusion does not hinge on a lucky choice of ``error''; we did not repeat the same-budget test for all $18$.

Two complications warrant control.
First, a kernel-noise floor: a control cell that passes through the merge mask with every cluster size equal to one has a $\log 1 {=} 0$ bias and should therefore reproduce the un-merged video exactly, yet a small residual LPIPS remains. We attribute this residual to nondeterminism and kernel/backend differences along the merge path rather than treating it as a single global measurement floor; we align the attention backend across arms and read correlations only where the measured differences exceed this residual.
Second, saturation: under very aggressive merging LPIPS stops discriminating between configurations, so we read correlations only in the regime where LPIPS remains informative.
Neither complication alters the conclusion that attention-output error is strongly correlated with video LPIPS on the evaluated budgets and candidate maps, making it a usable offline proxy for the allocation search.


\section{Additional ablations and comparisons}
\label{app:baselines}

All baselines share the locked budget $B{=}2880$ ($R{=}8$); every comparison therefore redistributes the same KV
tokens rather than spending more cache.

\paragraph{Candidate counts.} The distortion table is measured on a grid of $17$ representative counts
$\mathcal{R}_{\mathrm{meas}}=\{2,3,\dots,12,14,16,20,24,32,41\}$, denser near the uniform $r{=}8$. The main-table
allocators search the sub-grid $\mathcal{R}_{\mathrm{search}}=\{r\in\mathcal{R}_{\mathrm{meas}}:r_{\min}\le r\le
r_{\max}\}$ with $r_{\min}{=}6$ and $r_{\max}{=}24$. The cap binds---\method\ reaches $r{=}24$---so $32$ and $41$ are
measured but unused at this budget; whether relaxing the cap would help is left open. Continuous or off-grid proposals
(Sensitivity, Random) are rounded onto $\mathcal{R}_{\mathrm{search}}$ and repaired to $\sum P{=}B$.

\paragraph{Allocation baselines (merge operator fixed to ours).}
\begin{itemize}[leftmargin=1.4em,itemsep=1pt,topsep=2pt]
\item \textbf{Uniform.} $P_{l,h}{=}8$ everywhere.
\item \textbf{Random.} From uniform, apply budget-preserving $\pm1$ grid swaps (snapped to
$\mathcal{R}_{\mathrm{search}}$ and repaired to budget) to produce a mixed random reallocation, averaged over random seeds.
\item \textbf{Layer-only.} All heads in a layer share one count $R_l$, obtained by the same fixed-budget DP over the
$30$ head-averaged curves $\bar D_l(r)=\frac1H\sum_h D_{l,h}(r)$, spending $\sum_l H R_l{=}B$.
\item \textbf{Head-only.} Symmetric: each head index shares one count $R_h$ across layers, from the DP over the $12$
layer-averaged curves.
\item \textbf{Sensitivity.} $P_{l,h}\propto D_{l,h}(8)$ (single-point distortion at the uniform count), renormalized to
$\sum P{=}B$ and snapped to the grid; a single-point heuristic that ignores curve shape.
\item \textbf{Greedy.} Start every head at $r_{\min}$ and repeatedly advance the head with the largest marginal
distortion drop \emph{per added representative} to its next grid point, until $B$ is spent; converges near the DP
optimum.
\item \textbf{\method\ / \method-warm.} \Cref{alg:calibrate,alg:allocate}; \method-warm additionally gates on rollout position
(\cref{app:warm}).
\end{itemize}

\paragraph{Merge operators (budget fixed to uniform $R{=}8$).}
Every operator merges each head's $T$ temporal tokens to $R$ representatives and keeps the same count-weighted mean
\emph{value} and $\log$-count attention bias; they differ in the grouping and, for the baselines, the \emph{key}
aggregation (ours keeps a representative key, the baselines average keys). \Cref{tab:operators} is thus a comparison
of full operators, not of grouping alone.
\begin{itemize}[leftmargin=1.4em,itemsep=1pt,topsep=2pt]
\item \textbf{FPS-based merging (ours).} Farthest-point sampling on $L_2$-normalized keys picks $R$ representatives; each token joins
its nearest.
\item \textbf{Average pooling.} Contiguous strided groups of $\lceil T/R\rceil$ over the actual input length $T$.
\item \textbf{Random merging.} A fixed random assignment of tokens to $R$ groups (ToMe's own control).
\item \textbf{ToMe.} Bipartite soft matching \citep{bolya2022tome}: tokens split into alternating source/destination
sets, each source scored against destinations by key similarity, and the highest-scoring source$\to$destination edges
merged into their destination until $R$ tokens remain.
\item \textbf{DPC-kNN.} Density-peak clustering \citep{zeng2022not}: each token's density $\rho$ is its $k$-NN key
similarity and $\delta$ its distance to the nearest higher-density token; the $R$ tokens with largest
$\rho\!\cdot\!\delta$ are the peaks, and every token joins its nearest peak.
\item \textbf{AsymRnR-style.} We port the reduction rule of AsymRnR \citep{sun2025asymrnr}: $R/2$ fixed evenly-spaced
destination anchors plus $R/2$ adaptively kept, redundancy-ranked source tokens. Its official code targets
diffusers video DiTs (CogVideoX, Mochi, HunyuanVideo) and does not run on the autoregressive Wan cache, so we
re-implement the rule in our pipeline and label it \emph{AsymRnR-style} to distinguish it from the official method.
\end{itemize}

\paragraph{Allocation-map transfer to another operator.} \emph{ToMe@\method} applies the ToMe grouping under the frozen \method\ budget $P$---each
head $(l,h)$ merges to $P_{l,h}$ ToMe representatives, surplus slots masked out---\emph{transferring} the
FPS-calibrated map without re-fitting it to ToMe. It tests whether this map transfers directly to another operator. It does not:
under the faithful ToMe operator the transferred map hurts both full-clip ($0.213$ vs.\ $0.188$) and tail ($0.358$ vs.\
$0.331$). We conclude only that this FPS-calibrated map does not transfer directly to ToMe; whether a ToMe-calibrated map would help is a separate question we do not address.

\subsection{Budget sweep: where calibration pays off}
\label{app:budget}
\begin{figure}[H]
    \centering
    \includegraphics[width=\linewidth]{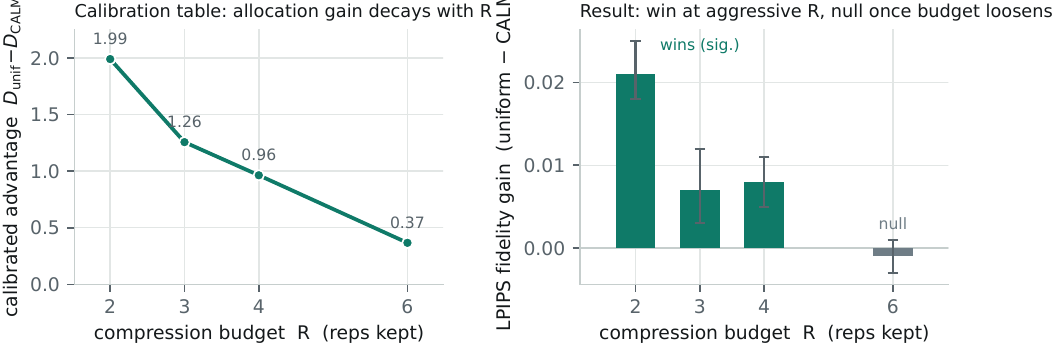}
    \caption{Comparison between calibrated attention error and video LPIPS across cache budgets on CausVid.
    The left panel shows the attention error reduction of \method\ relative to uniform allocation.
    The right panel shows the corresponding LPIPS reduction with confidence intervals.
    The proxy improves across all budgets, while the video quality benefit depends on compression strength.}
    \label{fig:mechanism}
\end{figure}

\begin{figure}[H]
\centering
\includegraphics[width=0.88\linewidth]{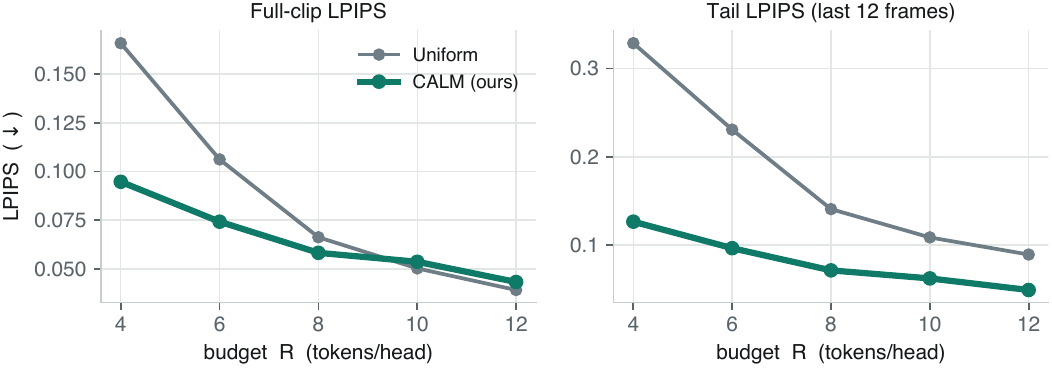}
\caption{Comparison of uniform and calibrated allocation across representative budgets.
The left panel shows full clip LPIPS, and the right panel shows LPIPS over the final temporal window.
Lower values indicate better quality.
The benefit of calibrated allocation is larger under tighter budgets and remains visible in later frames.}
\label{fig:budget}
\end{figure}

We sweep the per-head budget $R\in\{4,6,8,10,12\}$ for both uniform and \method\ allocation (merge operator fixed to
ours) on a $30$-prompt held-out subset (\cref{fig:budget}); because the sweep includes budgets as low as $R{=}4$, it
uses the grid-minimum floor $r_{\min}{=}2$ and the full measured grid up to $r_{\max}{=}41$, rather than the main table's $r_{\min}{=}6$ and $r_{\max}{=}24$; the $R{=}8$ point uses this same setting. \method's advantage is concentrated where it matters for
deployment. At tight budgets the full-clip gap is large ($R{=}4$: $0.095$ vs.\ $0.166$) and it closes as the cache
loosens; on the tail \method\ wins at \emph{every} budget, cutting the error by more than half at $R{=}4$ ($0.127$
vs.\ $0.329$). At loose budgets ($R{\ge}10$) uniform is already near-lossless on full-clip and the two differ by only a small mean gap, but the long-rollout benefit of calibration persists---consistent with the main-text finding that calibration
pays off in the constrained, long-rollout regime.

\subsection{\method-warm: temporal gating}
\label{app:warm}

\method-warm merges uniformly ($R{=}8$) while the rollout is inside the model's native $81$-pixel ($21$-latent)
window, then applies the calibrated map. The gate is keyed on the cumulative \emph{raw} latent-frame position, not the
compressed cache length---uniform merging pins the latter near $8$, so a ``compressed length'' trigger would never
fire. Merging happens at each block's end, after its clean KV is written: blocks whose start position is $<21$ merge
uniformly; the block at position $21$ (frames $21$--$23$) is still generated against uniform history, the first
calibrated merge fires immediately after it, and the first block that \emph{reads} the calibrated cache begins at
latent frame $24$. The early representatives are not frozen---they keep participating in later merges under the
calibrated quota---so \method-warm is a genuinely dynamic policy, not the global map applied from frame one.

\section{Cross-model evaluation details}
\label{app:generalization}

\subsection{Native calibration on each model}
The cross-model table in the main text (\cref{tab:crossmodel}) calibrates \method\ \emph{natively} on each generator: for every model we build a model-specific distortion table from its own dense rollouts, solve the budget-constrained dynamic program, freeze the resulting map, and compare it against uniform merging on that model's own held-out prompts. \Cref{tab:xmodel-protocol} summarizes the protocol.
CausVid \citep{yin2024causvid} uses the released distilled checkpoint with $48$-frame rollouts and budgets $R\in\{2,3,4,5,6\}$ evaluated on $30$ prompts. Causal Forcing \citep{zhu2026causalforcing} and the Self-Gradient-Forcing variant \citep{zhuang2026selfgradient} both use the framewise autoregressive architecture with a $21$-latent KV window and $4$ sink frames, $45$-frame rollouts, and budgets $R\in\{2,3,4,6\}$, and both are evaluated on the same $29$-prompt set, so $n{=}29$ for each. Causal Forcing uses the framewise \texttt{ar\_diffusion} initialization checkpoint, whereas SGF (marked $\dagger$ in \cref{tab:crossmodel}) uses the self-gradient-forcing-trained EMA checkpoint. Paired $95\%$ confidence intervals are bootstraps over the shared prompts.
With this initialization checkpoint the Causal Forcing uniform baseline is non-monotonic in $R$ ($0.186$, $0.186$, $0.212$, $0.196$ at $R{=}2,3,4,6$): raising the budget from $R{=}2$ to $R{=}4$ does not lower LPIPS, which cannot hold for a stable generator at fixed prompts and seed and instead reflects the higher generation variance of this checkpoint at $n{=}29$. We therefore do not treat the $R{=}2$ Causal Forcing comparison as a reliable regression, and read the better-behaved $R{=}4,6$ points---where uniform LPIPS decreases with budget as expected---as the more indicative result for this model.
For each target model we re-derive the candidate representative grid and bounds from its own KV history rather than copying Self-Forcing's; the three framewise generators (CausVid, Causal Forcing, and SGF) share the grid $\{1,2,3,4,6,8,11\}$ with $r_{\min}{=}1$ and $r_{\max}{=}11$. The budget is the summed representative count over the $30\times12$ units---$720$, $1080$, $1440$, $2160$, and $2880$ for $R\in\{2,3,4,6,8\}$, each equal to $R\times360$---while the sink, anchor, and recent frames are kept separately and are not counted against it. Because $r_{\min}{=}1$, an average of $R{=}2$ still admits a non-uniform map: the dynamic program drops some heads to a single representative and spends the freed budget on more sensitive ones, so the native $R{=}2$ maps range from $1$ up to $8$--$11$ representatives with mean exactly $2$; for $R\ge3$ a per-budget floor keeps the minimum from exceeding the average.

\begin{table}[H]
\centering
\small
\begin{tabular}{lccc}
\toprule
 & CausVid & Causal Forcing & SGF$^{\dagger}$ \\
\midrule
Checkpoint              & released distilled & framewise \texttt{ar\_diffusion} & framewise self-grad.\ (EMA) \\
Rollout (latent frames) & $48$               & $45$                            & $45$ \\
KV window / sink        & persistent         & $21$ / $4$                      & $21$ / $4$ \\
Budgets $R$             & $2$--$6$           & $2,3,4,6$                       & $2,3,4,6$ \\
Eval prompts $n$        & $30$               & $29$                            & $29$ \\
CI                      & paired bootstrap   & paired bootstrap                & paired bootstrap \\
\bottomrule
\end{tabular}
\caption{Native cross-model calibration protocol; results in \cref{tab:crossmodel}.
Each model is calibrated and evaluated on its own dense rollouts and held-out prompts with the same merge operator as the main experiments.
Causal Forcing and SGF share the framewise autoregressive architecture and the same $29$-prompt evaluation set (hence $n{=}29$ for both); \emph{SGF}$^{\dagger}$ denotes the self-gradient-forcing EMA-weight variant, distinct from the \texttt{ar\_diffusion} initialization used for Causal Forcing.}
\label{tab:xmodel-protocol}
\end{table}

\subsection{Transfer without target-model calibration}
Our main experiments evaluate Self-Forcing on the Wan2.1-T2V-1.3B backbone. To test whether the allocation gains
transfer beyond the calibrated generator, we port \method\ to \textbf{CausVid} \citep{yin2024causvid}---an
independently distilled autoregressive generator that shares the Wan backbone but a different training procedure and
persistent-history KV path. All comparison arms use the same mask-aware SDPA attention backend, including the
Dense reference; this reference differs from the released Flash-attention execution path.
On the tested prompt and seed, enabling the merge hook with a budget large enough to bypass reduction reproduces
this backend-matched Dense output exactly (measured MSE $0$). This checks the no-reduction path, rather than
equivalence to the released execution backend. We then compare Uniform with the transferred allocation at three
matched budgets on $30$ held-out prompts with matched noise and sampler settings.

Crucially this is a \emph{transfer}, not a re-calibration: we take the per-(layer,head) allocation \emph{ranking}
calibrated on Self-Forcing and water-fill it into the adapter's budget grid, holding the total representative budget
equal to Uniform's. The transferred allocations use a maximum quota of $11$ representatives per head; this is an
allocation constraint, not CausVid's native history length. No CausVid distortion table is used to construct these maps.

\begin{table}[H]
\centering
\begin{tabular}{cccccc}
\toprule
$R$ & budget $B$ & Uniform $\downarrow$ & \method $\downarrow$ & $\Delta$ (\method $-$ Uniform) & wins \\
\midrule
$8$ & $2880$ & $0.0100$ & $0.0105$ & $+0.0005$ \; [$-0.0002,+0.0011$] & $12/30$ \\
$4$ & $1440$ & $0.0204$ & $0.0212$ & $+0.0008$ \; [$-0.00003,+0.00192$] & $12/30$ \\
$2$ & $720$  & $0.0545$ & $\mathbf{0.0457}$ & $\mathbf{-0.0088}$ \; [$-0.0124,-0.0054$] & $\mathbf{28/30}$ \\
\bottomrule
\end{tabular}
\caption{Transfer of the Self Forcing allocation map to CausVid \textbf{without target model calibration}.
$R$ and $B$ specify the cache budget, $\Delta$ is the paired LPIPS difference from uniform allocation, and wins counts improved prompts.
Lower LPIPS is better.
The transferred map improves quality only under the strongest compression setting.}
\label{tab:generalization}
\end{table}

\Cref{tab:generalization} reports all three budgets. At $R{=}8,4$, Uniform incurs relatively low LPIPS
($0.0100$ and $0.0204$). Transfer yields slightly higher point estimates, with paired $95\%$ confidence intervals
that include zero and $12/30$ prompt wins at each budget. These results do not establish an improvement or
statistical equivalence. At $R{=}2$, Uniform's LPIPS rises to $0.0545$, and the transferred map reduces it to
$0.0457$ (approximately $16\%$), winning $28/30$ prompts; the paired confidence interval excludes zero.
Thus, transfer helps at the tightest tested budget, supporting the usefulness of the Self-Forcing allocation ranking
on this second generator without target-model calibration.

Equal representative counts need not produce equal distortion across generators and rollout lengths.
For example, Uniform at $R{=}8$ has LPIPS $0.149$ on our Self-Forcing evaluation but $0.0100$ on CausVid.
This motivates interpreting gains alongside each baseline's distortion, rather than assuming that equal $R$ defines
equal compression difficulty. These experiments do not match distortion across models: even CausVid at $R{=}2$
has lower baseline LPIPS than Self-Forcing at $R{=}8$, and the rollout protocols differ.
The same caveat applies within CausVid: this subsection's transferred-map Uniform LPIPS ($0.0545$ at $R{=}2$) is lower
than the native CausVid Uniform in \cref{tab:crossmodel} ($0.135$ at $R{=}2$). The two use different rollout protocols,
so their absolute LPIPS are not directly comparable across the two tables.
Low baseline error limits the available absolute reduction but does not rule out a gain at a less aggressive budget.
The observed budget dependence is empirical evidence, not a proof that calibration can help only in a constrained regime.

The \emph{transfer} evidence here is limited to CausVid ($30$ prompts, with a positive result at $R{=}2$); native
calibration on three generators is reported separately in \cref{tab:crossmodel}. Transferring a source allocation ranking and recalibrating on the target model answer
different questions. The former tests whether the allocation pattern transfers; the latter requires building a
CausVid-native distortion table and solving its budget-constrained allocation problem.
A comparison of Uniform, the transferred map, and a native map under the same target-model protocol would assess
both; native calibration results (including CausVid) are reported in \cref{tab:crossmodel} and \cref{tab:xmodel-protocol}.
Even so, native calibration is not guaranteed to improve video LPIPS over transfer, because dynamic programming
minimizes the calibrated additive proxy within its candidate set, rather than end-to-end video distortion.

Full-sequence diffusion requires a separate extension: compressing the K/V used by each attention call differs from
compressing the accumulated history of an autoregressive generator. Such an extension needs a model-specific adapter,
an explicit grouping domain and budget definition, and independent validation. Transfer across backbone architectures
and this full-sequence extension remain future work.

\section{Additional VBench results}
\label{app:vbench-full}
\begin{table}[H]
\centering
\begin{tabular}{lcccc}
\toprule
Dimension $\uparrow$ & $n$ & Dense & Uniform & \method \\
\midrule
Subject Consistency    & 72  & 0.948 & 0.928 & \textbf{0.941} \\
Background Consistency  & 85  & 0.955 & 0.947 & \textbf{0.952} \\
Temporal Flickering    & 75  & 0.990 & 0.989 & \textbf{0.990} \\
Motion Smoothness      & 72  & 0.984 & 0.983 & \textbf{0.984} \\
Imaging Quality        & 92  & 0.698 & 0.688 & \textbf{0.697} \\
Aesthetic Quality      & 92  & 0.628 & 0.613 & \textbf{0.622} \\
\midrule
Dynamic Degree         & 72  & 0.653 & 0.611 & 0.597          \\
Human Action           & 100 & 0.810 & 0.810 & 0.810          \\
Appearance Style       & 90  & 0.199 & 0.202 & 0.200          \\
Temporal Style         & 94  & 0.231 & 0.233 & 0.232          \\
Overall Consistency    & 92  & 0.241 & 0.242 & 0.241          \\
\bottomrule
\end{tabular}
\caption{VBench comparison across all available scoreable dimensions.
$n$ is the number of matched samples, and higher scores indicate better quality.
The unavailable object detection dimensions are omitted.
\method\ matches or outperforms uniform allocation on the quality and consistency dimensions, while the other dimensions remain similar.}
\label{tab:vbench-full}
\end{table}

For completeness we report every VBench dimension whose scorer was available to us---eleven of the sixteen; the five
GRIT-based object-detection dimensions are omitted because the VBench GRIT checkpoint was not available in this
evaluation (\cref{tab:vbench-full}). Beyond the six compression-relevant dimensions of \cref{tab:vbench}, the style and
alignment dimensions (Human Action, Appearance Style, Temporal Style, Overall Consistency) change little under
compression in this evaluation: the three arms differ only by small margins. Dynamic Degree is reported but read with
care: VBench itself notes the metric is easily gamed---it scores raw motion magnitude, not fidelity, so a lower value
is not necessarily better or worse. Here \method\ ($0.597$) and uniform ($0.611$) both fall below Dense ($0.653$); we do
not read this as a quality difference on its own, and it should be interpreted alongside the consistency and fidelity
dimensions.

\section{Efficiency and calibration cost}
\label{app:efficiency}

\begin{table}[H]
\centering
\small
\begin{tabular}{lccc}
\toprule
 & Dense & Uniform ($R{=}8$) & \method \\
\midrule
Theoretical logical KV (GB) $\downarrow$ & $12.9$ & $2.3$  & $2.3$ \\
Realized KV buffer (GB)                   & $12.9$ & $12.9$ & $12.9$ \\
Peak GPU memory (GB)                      & $30.0$ & $30.1$ & $30.1$ \\
Per-video generation time (s)             & $31.0$ & $30.2$ & $34.7$ \\
Offline calibration (one-time)            & ---    & ---    & $50\times17$ shadow-probed rollouts \\
\bottomrule
\end{tabular}
\caption{Efficiency accounting, separating three distinct memory quantities.
\emph{Theoretical logical KV} is the ideal compact cache at the representative budget ($B{=}2880$ for the compressed arms); \emph{realized KV buffer} is the resident cache our implementation actually holds (the full $45$-frame buffer, merged slots marked dead rather than freed), so it is identical across methods; \emph{peak GPU memory} is the measured device peak.
Peak memory and per-video generation time are measured on a single RTX PRO 6000 Blackwell over the evaluation prompts (per-video means; \method's median is $35.6$\,s).
The frozen allocation adds no online allocation search at inference (the dynamic program runs once, offline). In this measured setting \method\ runs about $15\%$ slower than Uniform per video; the per-head padding of non-uniform maps (each layer's heads padded to its widest head) may contribute this overhead, which we have not profiled further.
Calibration is a one-time offline cost of $50\times17$ dense rollouts, each instrumented with shadow-cache probing, plus a negligible dynamic program.}
\label{tab:efficiency}
\end{table}

We keep three quantities distinct (\cref{tab:efficiency}). \emph{(i) Logical cache.} At budget $B$ a compressed arm holds
$\sum_{l,h}P_{l,h}{=}B$ temporal representatives; each representative is one key/value over $S{=}1560$ spatial
positions per frame, so the ideal compact size is $2\,b\,S\,d_h\,q\,B$ bytes (K and V; batch $b$, $q$ bytes/element).
At $B{=}2880$ this is $\approx\!2.3$\,GB versus Dense's $\approx\!12.9$\,GB, the $45/8\approx5.6\times$ ratio---a
\emph{theoretical} bound. \emph{(ii) Realized cache.} Our implementation pre-allocates the KV buffer for the full
$45$-frame rollout and marks merged slots dead rather than freeing them, and heads are padded to their layer's widest
head; the \emph{resident} buffer is therefore $\approx\!12.9$\,GB for every method regardless of $B$. \emph{(iii)
Latency.} The frozen allocation adds no online allocation search at inference---the dynamic program runs once,
offline. In this measured setting \method\ generates video about $15\%$ slower than uniform in our current
implementation (per-video means $30.2$ vs.\ $34.7$\,s; \cref{tab:efficiency}). A non-uniform map pads each layer's
heads to its widest head, which may contribute this overhead; we have not profiled it further. We accordingly report
compression as a logical/theoretical quantity: the paper's claim is \emph{quality at a fixed representative budget},
not a realized memory or speed win. Whether compact storage and a matching ragged or compaction-aware attention kernel
recover the $5.6\times$ and remove this overhead is left to verify.

\end{document}